\pdfoutput=1

\ifdefined\REVIEW
    \documentclass[
        a4paper,
    ]{article}
    \usepackage{lineno, setspace}
    \modulolinenumbers[5]
    \pagewiselinenumbers
\else
    \documentclass[
        a4paper,
        twocolumn
    ]{article}
\fi

\usepackage{IEK10} \usepackage{natbib}

\def\IEK10{
  Forschungszentrum Jülich GmbH,
  Institute of Energy and Climate Research,
  Energy Systems Engineering (IEK-10),
  Jülich 52428,
  Germany
}
\def\JARA{
  JARA-ENERGY, 
  Jülich 52425,
  Germany
}
\def\SVT{
  RWTH Aachen University,
  Process Systems Engineering (AVT.SVT),
  Aachen 52074,
  Germany
}
\def\IEKSTE{
  Forschungszentrum Jülich GmbH,
  Institute of Energy and Climate Research,
  Systems Analysis and Technology Evaluation (IEK-STE),
  Jülich 52428,
  Germany
}
\def\Colone{
    Institute for Theoretical Physics, 
    University of Cologne, 
    50937 K\"oln, 
    Germany
}

\newcommand{\mytitle}{
Multivariate Probabilistic Forecasting of Intraday Electricity Prices using Normalizing Flows
}

\newcommand{\affil}{
  \begin{itemize}[leftmargin=3mm, itemsep=0mm]
    \item[$^a$]\IEK10
    \item[$^b$]\SVT
    \item[$^c$]\IEKSTE
    \item[$^d$]\Colone
    \item[$^e$]\JARA
\end{itemize}
}

\def\firstAuthor{Eike Cramer}
\newcommand{\myauthor}{
\firstAuthor$^{a,b}$\orcidlink{0000-0002-6882-5469}, 
Dirk Witthaut$^{c,d}$\orcidlink{0000-0002-3623-5341},
Alexander Mitsos$^{e,a,b}$\orcidlink{0000-0003-0335-6566}, 
Manuel Dahmen$^{a,*}$\orcidlink{0000-0003-2757-5253} }
\newcommand{\correspondingauthor}{Manuel Dahmen, \IEK10 \newline E-mail: m.dahmen@fz-juelich.de}
\author{\myauthor}

\usepackage[
  colorlinks,
  linkcolor=blue,
  citecolor=blue,
  urlcolor=blue,
  pdftitle={\mytitle},
  pdfauthor={\firstAuthor}
]{hyperref}
\usepackage[capitalise, nameinlink]{cleveref}
\crefname{table}{Tab.}{Tab.}

\usepackage{orcidlink}

\newcommand{\setpgfexternalcounter}[1]{
  \makeatletter \pgfkeysgetvalue{/tikz/external/figure name}\myexternalname
  \expandafter\gdef\csname c@tikzext@no@\myexternalname\endcsname{#1}\makeatother
}

\begin{document}

\ifx\REVIEW\undefined
\twocolumn[
\begin{@twocolumnfalse}
\fi
  \thispagestyle{firststyle}

  \begin{center}
    \begin{large}
      \textbf{\mytitle}
    \end{large} \\
    \myauthor
  \end{center}

  \vspace{0.5cm}

  \begin{footnotesize}
    \affil
  \end{footnotesize}

  \vspace{0.5cm}

Electricity is traded on various markets with different time horizons and regulations. Short-term intraday trading becomes increasingly important due to the higher penetration of renewables. 
In Germany, the intraday electricity price typically fluctuates around the day-ahead price of the European Power EXchange (EPEX) spot markets in a distinct hourly pattern. 
This work proposes a probabilistic modeling approach that models the intraday price difference to the day-ahead contracts. 
The model captures the emerging hourly pattern by considering the four 15\,min intervals in each day-ahead price interval as a four-dimensional joint probability distribution. 
The resulting nontrivial, multivariate price difference distribution is learned using a normalizing flow, i.e., a deep generative model that combines conditional multivariate density estimation and probabilistic regression.
Furthermore, this work discusses the influence of different external impact factors based on literature insights and impact analysis using explainable artificial intelligence (XAI). 
The normalizing flow is compared to an informed selection of historical data and probabilistic forecasts using a Gaussian copula and a Gaussian regression model.
Among the different models, the normalizing flow identifies the trends with the highest accuracy and has the narrowest prediction intervals. 
Both the XAI analysis and the empirical experiments highlight that the immediate history of the price difference realization and the increments of the day-ahead price have the most substantial impact on the price difference.  
\noindent
\\
\textbf{Keywords:}
Electricity price forecasting;
probabilistic forecasting;
deep learning;
multivariate modeling 
  \vspace*{5mm}
\ifx\REVIEW\undefined
\end{@twocolumnfalse}
]
\fi

  \newpage

  \section{Introduction}\label{sec:Price_Delta_Introduction}
The liberalization of the European electricity markets has introduced the auction-based day-ahead market and the continuously traded intraday market \citep{mayer2018electricity}. 
The hourly day-ahead auction market is cleared on the day before delivery. Afterward, the continuous 15\,min interval intraday markets aim to reduce the balance requirements by enabling short-term changes to delivery and consumption commitments and react to the ramping of supply and demand \citep{ocker2017german, EPEX2021Documentation, BUBLITZ20191059}. 
The increasing penetration of renewable electricity sources with low marginal cost \citep{sensfuss2008merit} causes intraday prices to become more volatile \citep{viehmann2017state,shinde2019literature}. 
Thus, electricity price forecasting (EPF) \citep{weron2014electricity} for the intraday market becomes increasingly difficult \citep{lago2021forecasting, jedrzejewski2022electricity} as the forecasts have to account for the inherent uncertainty of intermittent renewable electricity supply and human interaction on the continuous market \citep{mayer2018electricity, BUBLITZ20191059}. Still, accurate forecasts of intraday prices can lead to significant financial gains \citep{SERAFIN2022106125}.

A promising approach to address the uncertainty in EPF uses probabilistic forecasting models that predict a distribution of possible realizations instead of a single point forecast \citep{zhang2003energy, Nowotarski2018Recent}. 
The quantitative knowledge of the uncertainty allows the market participants to develop different strategies considering possible outcomes and to derive more effective bidding strategies \citep{Nowotarski2018Recent}.
Established approaches for probabilistic forecasting include prediction intervals \citep{zhang2003energy,serinaldi2011distributional,jonsson2014predictive,uniejewski2021regularized} and density forecasts \citep{huurman2012power,panagiotelis2008bayesian,andrade2017probabilistic,narajewski2020ensemble}. 
Other works use ensemble forecasts \citep{narajewski2020ensemble} or propose combinations of deterministic and probabilistic approaches \citep{marcjasz2020probabilistic}. There is also an increasing trend towards machine learning methods \citep{jkedrzejewski2022electricity}.

Most probabilistic EPF papers utilize univariate modeling approaches \citep{Nowotarski2018Recent}. Here, the models predict single time steps in a step-by-step fashion and consider the correlation between time steps via the autoregressive components. Alternatively, multivariate models predict sequences in multi-step forecasting \citep{panagiotelis2008bayesian,ziel2018day}. For instance, \cite{panagiotelis2008bayesian} use a multivariate student-t distribution model for intraday prices in Australia. 
In a study on deterministic forecasting models, \cite{ziel2018day} report on the advantages of using a multivariate approach for day-ahead EPF. 
Most of the literature on probabilistic EPF only considers the day-ahead markets \citep{lago2021forecasting}. A notable exception is presented by \cite{andrade2017probabilistic}, who perform probabilistic forecasting for both day-ahead and intraday markets.

The literature on intraday EPF generally addresses the intraday prices as a stand-alone prediction problem without considering the relation to the day-ahead prices \citep{pape2016fundamentals, kremer2021econometric}. 
However, separating the intraday EPF from the day-ahead prices ignores the connection between the two markets, where the intraday market mainly serves to make adjustments to the fixed day-ahead contracts \citep{koch2019short}.
Consequently, the intraday price follows the day-ahead price trend.
Furthermore, the adjustments react to the ramping of supply and demand, which leads to steadily increasing or decreasing prices in the four intraday trading intervals of each day-ahead trading interval \citep{kiesel2017econometric}.
Thus, there is a distinct hourly fluctuation of the intraday prices around the day-ahead prices that averages to the day-ahead price \citep{han2022complexity}.

This work proposes a modeling approach for intraday electricity prices that exploits the special relationship between the two electricity markets. 
Assuming that the intraday EPF is performed short-term, i.e., after the day-ahead price is set, the intraday price is modeled via the difference between the two market prices. 
Furthermore, the hourly increasing or decreasing patterns of the four 15\,min intervals of each day-ahead price interval are captured using a multivariate forecasting scheme. 
Hence, the forecasting problem needs to predict the four-dimensional price difference vector probability distribution which is defined by heavy tails, high correlations between the dimensions, and conditional dependence on external impact factors. Thus, the price difference vector distribution cannot be modeled via standard probability distribution models. 
Instead, this work uses normalizing flows \citep{papamakarios2021normalizing}, a non-parametric, i.e., fully data-driven, distribution model, to learn the probability distribution of the price difference vector. 
Normalizing flows learn complex distributions without a priori assumptions about the data and can include external impact factors for probabilistic regression \citep{winkler2019learning, rasul2021multivariate}. 
For scenario generation of other energy time series, such as renewable electricity generation and electricity demand, normalizing flows have already shown promising results \citep{dumas2021deep, cramer2022pricipalcomponentflow, cramer2022conditional, arpogaus2021probabilistic}.
The forecasting performance of the normalizing flow is compared to an informed selection of historical data, a Gaussian copula, and a multivariate Gaussian regression.

Both day-ahead and intraday prices are heavily influenced by external impact factors such as renewable electricity generation, import and export, and demand \citep{trebbien2022understanding}. However, the literature is scarce on evaluations on the impact factors for the price \textit{differences} analyzed and modeled in this study. 
Thus, this work investigates the impact of different external factors on the realization of the price difference vector using explainable artificial intelligence (XAI).
XAI is a branch of machine-learning research aiming to design human-understandable models and post-modeling explanations for black-box models \citep{gunning2017explainable, tjoa2020survey}.
An important subbranch of post-modeling explanations in XAI considers the predictability of the desired labels or values given the considered input features \citep{lundberg2020local}. 
In particular, this work uses SHAP-values (SHapley Additive exPlanations), which are a typical approach to quantify the impact of each input feature on the output of the machine-learning model. 
In previous works, SHAP-values have been successfully applied to analyze the predictability of power grid frequencies \citep{kruse2020predictability}, to reveal drivers and risks for power grid frequency stability \citep{kruse2021revealing}, predict the electricity mix in the African grid \citep{alova2021machine}, and to gain insight into solar power forecasting \citep{kuzlu2020gaining}.

The remainder of this work is organized as follows: 
Section~\ref{sec:Price_Delta_Methodology_4D_modeling} introduces the four-dimensional price difference modeling approach and highlights the advantages of multivariate modeling over the traditional univariate approach. 
Section~\ref{sec:Price_Delta_DistributionModeling} introduces the different multivariate probabilistic models.
Section~\ref{sec:Price_Delta_InputFeatureSelection} discusses relevant external input factors and performs the feature selection using SHAP-values analysis.  
Section~\ref{sec:Price_Delta_Results} applies the different distribution modeling approaches to learn the four-dimensional conditional distribution and discusses the impact of each external input factor via an empirical study. 
Finally, Section~\ref{sec:Price_Delta_Conclusion} concludes this work.   \section{Electricity price analysis}\label{sec:Price_Delta_Methodology_4D_modeling}
This Section first describes the used data sets. Second, an approach for multivariate modeling of intraday electricity prices is proposed. Finally, the Section compares a univariate and a multivariate sampling approach for the price difference vector using assessment scores for multivariate probabilistic forecasting.

\subsection{Data description}
The majority of electricity is traded on the auction-based day-ahead market, while the intraday market is primarily used to make short-term adjustments to the previously submitted bids \citep{kiesel2017econometric, koch2019short, han2022complexity}. 
The intraday contracts are negotiated between the producers and the consumers on different time horizons and for different periods. Thus, different prices occur for the same intraday trading interval \citep{mayer2018electricity,ocker2017german}.
As an indicator of the intraday electricity prices, the market operators release indices of volume-weighted averages \citep{EPEX2021Documentation}. 
This work uses the ID$_3$ index, i.e., the index reflecting the volume-weighted average price of the last three hours before delivery, as it is the most commonly used indicator \citep{EPEX2021Documentation, BUBLITZ20191059}.
The data set considered in this work is the day-ahead and ID$_3$ price time series recorded over the years of 2018 and 2019 provided by the \cite{EnergyCharts2022}. 
The day-ahead forecasts and actual values for the renewable electricity production and the loads are taken from \cite{entsoe2022}.
All renewable electricity time series are recorded in 15\,min intervals.
Note that the electricity price data from 2020 and 2021 shows atypical behavior due to the COVID-19 pandemic \citep{narajewski2020changes, badesa2021ancillary} and is, therefore, not included. 

\subsection{Hourly price difference vectors}\label{sec:Price_Delta_HourlyPriceDifference}
The volume of electricity traded on the day-ahead and the intraday markets exemplifies the relationship between the two markets. In 2019, the volume of day-ahead trades was 501.6\,TWh, and the intraday trades amounted to 83.2\,TWh \citep{EPEX2019TransparencyInternational}. These numbers highlight how traders primarily trade on the day-ahead market and then adjust their bids on the intraday market.
Intraday trades address ramping of supply and demand within the day-ahead trading intervals and compensate for forecast errors of demand and renewable electricity supply \citep{kiesel2017econometric,spodniak2021impact}. 
In particular, ramping of large suppliers and consumers leads to a distinct pattern of increasing or decreasing intraday prices within each day-ahead trading interval \citep{kiesel2017econometric}. 
For instance, a scheduled ramp-up of electricity supply results in an undersupply in the first 15 minutes and an oversupply in the last 15 minutes of the respective hour, which leads to higher prices in the first 15 minutes and lower prices in the last 15 minutes, respectively. 
Thus, the hour-wise constant day-ahead contracts lead to hourly patterns in the 15\,min intraday prices \citep{kiesel2017econometric, han2022complexity}.

\begin{figure}
    \centering
    \includegraphics[width=\columnwidth]{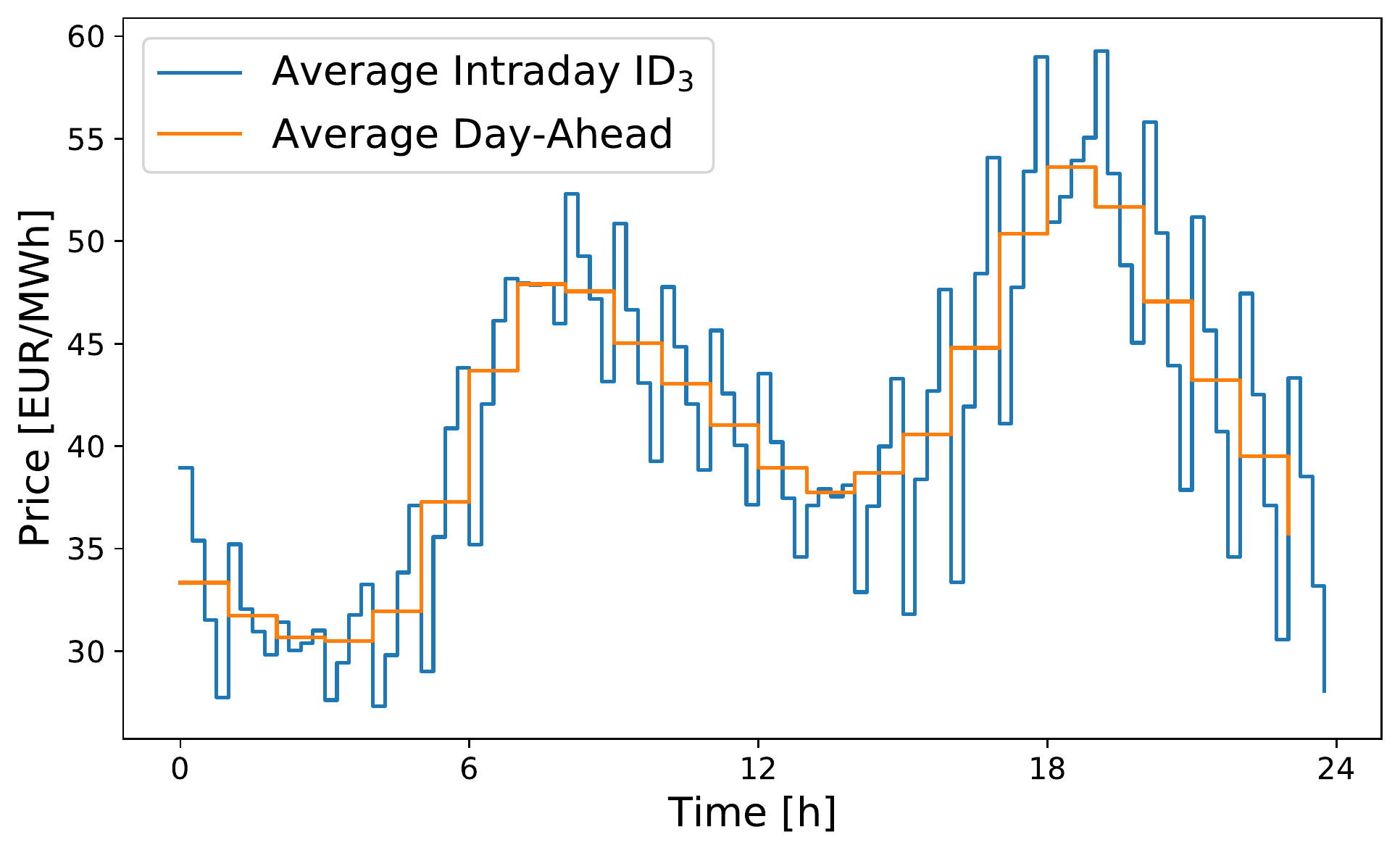}
    \caption{Average daily profiles of day-ahead and ID$_3$ price trends of 2018 and 2019. EPEX spot price data from \cite{EnergyCharts2022}.}
    \label{fig:Price_Delta_AverageDay}
\end{figure}
Figure~\ref{fig:Price_Delta_AverageDay} shows the average daily day-ahead and intraday prices in 2018 and 2019, i.e., each step represents the average price interval over the two years. 
As indicated by \cite{kiesel2017econometric}, the average ID$_3$ price within each hour is close to the respective day-ahead price. Furthermore, the ID$_3$ price either increases or decreases within the respective day-ahead trading intervals. 
These observations lead to the following conclusions about the relationship of the two price time series:
First, the ID$_3$ index fluctuates around the day-ahead auction price.
Second, Figure~\ref{fig:Price_Delta_AverageDay} highlights that the distinct hourly patterns of either increasing or decreasing steps appear systematically. 
Exceptions of this pattern only appear at local minima and maxima of the day-ahead price.
In summary, the relationship between the day-ahead and the intraday markets as primary and adjustment markets leads to a distinct hourly pattern of intraday price fluctuations around the day-ahead prices. 

The proposed modeling-approach for the ID$_3$ price index aims to capture the observed patterns. 
Since the intraday market is used for adjustments to the day-ahead market, the difference between the two price time series represents the characteristics of the intraday market.
Assuming that the prediction takes place after the day-ahead auction has been settled and the day-ahead prices are known, the ID$_3$ price index can be described via the sum of the day-ahead price and the difference between the two price time series. 
Furthermore, the hourly pattern of fluctuations is captured by modeling the four steps of price differences in each day-ahead trading interval as a four-dimensional vector. 
For a given hour $t$ the price difference vector $\bm{\Delta}\textbf{ID}_3$ then reads:
\begin{equation} \label{Eq:Price_Delta_Definition_Price_Difference_vector}
    \bm{\Delta}\textbf{ID}_3(t)
    = 
    \left[\begin{matrix}
        \text{ID}^{00}_3(t) - \text{DA}(t) \\
        \text{ID}^{15}_3(t) - \text{DA}(t) \\
        \text{ID}^{30}_3(t) - \text{DA}(t) \\
        \text{ID}^{45}_3(t) - \text{DA}(t) 
    \end{matrix}\right] 
\end{equation}
The four dimensions each describe the difference between intraday (ID$_3^{XX}(t)$) and day-ahead (DA$(t)$) price for a 15\,min interval in the hour $t$. The superscripts $00, 15, 30, \text{and}~45$ indicate the starting minutes of the four intraday trading intervals. 

Note that while the day-ahead market sets the prices for all time intervals simultaneously, the intraday market allows for trading up to five minutes prior to the delivery, and the prices for each trading interval are, thus, set independently. 
Intuitively, a step-by-step forecasting approach, where each step is informed by the previous, could be considered more appropriate than the chosen multivariate approach. 
However, the multivariate modeling approach is specifically designed to consider the correlation between the forecast dimensions, i.e., the four quarter-hour intervals, and there are no limitations to the practical application of the multivariate modeling approach. 

\subsection{Analysis of the price difference vector distribution}
To showcase the advantage of the multivariate modeling approach, this Subsection first highlights the strong correlation between the four price difference time steps in each day-ahead trading interval by showing the Pearson correlation and then compares a univariate and a multivariate sampling approach to highlight how the multivariate approach better captures the correlations between the time steps. 

Table~\ref{tab:Price_Delta_Pearson_PriceDifference} shows the Pearson correlation of the four dimensions in the joint distribution of price differences, i.e., the autocorrelations of the respective time steps. 
The correlations between neighboring time steps are high and, thus, highlight the strong correlation between the time steps within each day-ahead trading interval. In fact, it appears that each interval is predetermined by the previous dimension, which motivates the multivariate modeling approach. 
\begin{table}
    \centering
    \caption{Pearson correlation within the 4D-$\bm{\Delta}\textbf{ID}_3$ distribution (see Equation~\eqref{Eq:Price_Delta_Definition_Price_Difference_PDF}). Price data from \cite{EnergyCharts2022}.}
    \label{tab:Price_Delta_Pearson_PriceDifference}
\begin{tabular}{@{}lllll@{}}
    \toprule
            & $\Delta\text{ID}_3^{00}$ & $\Delta\text{ID}_3^{15}$ & $\Delta\text{ID}_3^{30}$ & $\Delta\text{ID}_3^{45}$ \\ \midrule
    $\Delta\text{ID}_3^{00}$ & 1.00    & 0.77    & 0.41    & 0.09    \\
    $\Delta\text{ID}_3^{15}$ & 0.77    & 1.00    & 0.74    & 0.46    \\
    $\Delta\text{ID}_3^{30}$ & 0.41    & 0.74    & 1.00    & 0.83    \\
    $\Delta\text{ID}_3^{45}$ & 0.09    & 0.46    & 0.83    & 1.00    \\ \bottomrule
    \end{tabular}\end{table}

To confirm that the multivariate approach captures the correlation of neighboring time steps better than the classical univariate approach, two different approaches of univariate and multivariate selection of historical data are compared using scoring metrics for multivariate probabilistic forecasts.
Figure~\ref{fig:Price_Delta_AverageDay} shows a strong dependency of the average realization on the hour of the day.
The multivariate selection is based on this observation and randomly selects historical realizations of the hourly price difference intervals that occurred during the same hour of the day. 
Similarly, the univariate selection approach randomly selects values that occurred during the same 15\,min interval and combines the selections into four-dimensional samples. 
Both approaches randomly select price differences from January 2018 to June 2019. July 2019 is set aside as a test set.

Unlike point forecasts, probabilistic forecasts cannot be evaluated by residual metrics such as the mean squared error or the mean absolute error. 
Instead, probabilistic forecasts are typically evaluated in terms of reliability and sharpness \citep{Nowotarski2018Recent}. Here, reliability describes whether the realization lies within certain prediction intervals.
Sharpness analyzes the tightness of the prediction intervals, i.e., how close the lower and upper boundaries of the prediction intervals are together \citep{Nowotarski2018Recent}. 
A good probabilistic forecast should be both reliable and sharp to avoid missing the realization and avoid large prediction intervals that carry little information. 

To evaluate the different sampling approaches, this work uses the energy score  \citep{gneiting2007strictly,pinson2012evaluating} and the variogram score \citep{scheuerer2015variogram}. 
The energy score \citep{gneiting2008assessing,pinson2012evaluating} is a widely applied metric for multistep probabilistic forecasts. It assesses how close the prediction samples are to the realizations and how diverse the samples are:
\begin{equation*}\label{Eq:Price_Delta_DefinitionEnergyScore}
    \text{ES} = 
        \frac{1}{N} \sum_{s=1}^{N} \vert\vert \mathbf{x} - \hat{\mathbf{x}}_s \vert\vert_2
        - \frac{1}{2{N}^2} \sum_{s=1}^{N} \sum_{s'=1}^{N} \vert\vert \hat{\mathbf{x}}_s - \hat{\mathbf{x}}_{s'} \vert\vert_2
\end{equation*}
Here, $\mathbf{x}$ is the realized ID$_3$ price-vector, i.e., the actual realization from the test set, $\hat{\mathbf{x}}_s$ are sample vectors drawn from the predicted distributions, $N$ is the number of samples, and $\vert\vert \cdot \vert\vert_2$ is the 2-norm.
The variogram score \citep{scheuerer2015variogram} aims to assess whether a multivariate forecast correctly describes the correlation between the individual time steps:
\begin{equation*}
    \text{VS} =\frac{1}{N} \sum_{t=1}^{\hat{T}}\sum_{t^{'}=1}^{\hat{T}} \left( \vert x_{t} - x_{t^{'}} \vert^{\gamma}
            - \frac{1}{N} \sum_{s=1}^{N} \vert \hat{x}_{t}^{s} - \hat{x}_{t^{'}}^{s} \vert ^{\gamma} \right)^{2}
\end{equation*}
Here, $x_{t}$ are the realized ID$_3$ price time steps, $\hat{x}_{t}$ are time steps of samples from the predicted ID$_3$ distribution, $\hat{T}=4$ is the dimension of $x$, i.e., the number of 15\,min intervals in the multivariate prediction, $N$ is the number of samples, $\gamma$ is the variogram order (typically $\gamma=0.5$ \citep{scheuerer2015variogram}), and $\vert \cdot \vert$ denotes the absolute value. 
Both energy score and variogram score are negatively oriented metrics, i.e., lower values indicate better performance \citep{scheuerer2015variogram, pinson2012evaluating}.

\begin{figure}
    \centering
    \includegraphics[width=\columnwidth]{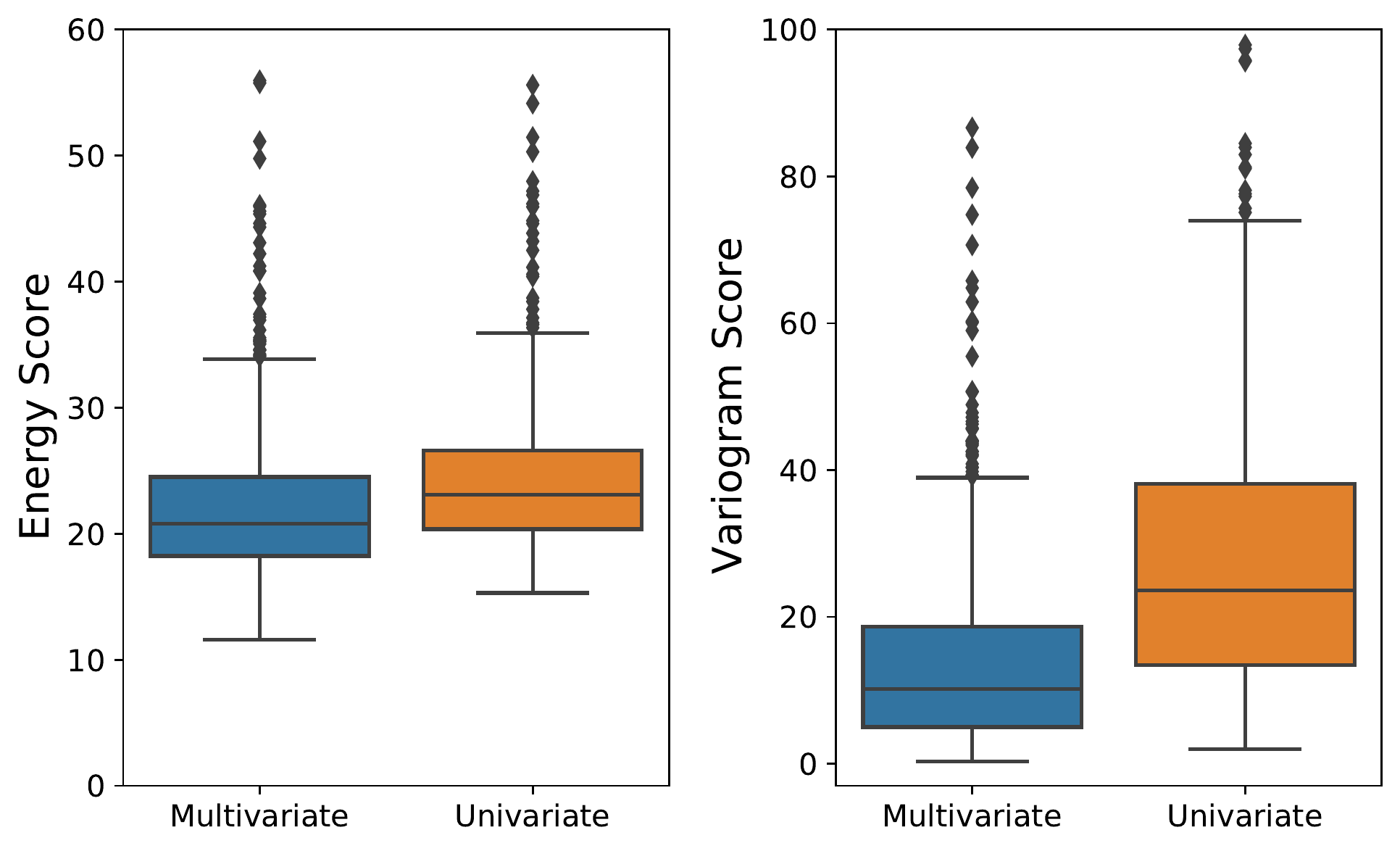}
    \caption{Box plots \citep{waskom2021seaborn} of the energy score \citep{gneiting2007strictly,pinson2012evaluating} and variogram score \citep{scheuerer2015variogram} distributions of historical data selected using univariate and multivariate approaches. 
    Price data from January 2018 to June 2019 \citep{EnergyCharts2022} is used and the evaluation is performed on July 2019 as the test set.}
    \label{fig:Price_Delta_Scores_Uni_Mulit}
\end{figure}
Figure~\ref{fig:Price_Delta_Scores_Uni_Mulit} shows box plots \citep{waskom2021seaborn} of the energy score and the variogram score distributions of the univariate and the multivariate selection approaches applied to each day-ahead trading interval of the test month, respectively.
The multivariate approach yields lower values for energy scores and variogram scores and, thus, indicates better agreement with the realizations. 
In particular, the variogram score suggests that the univariate selection fails to represent the correct correlation between the time steps whereas the multivariate approach captures the hourly pattern. 
In conclusion, the results in Figure~\ref{fig:Price_Delta_Scores_Uni_Mulit} confirm that the multivariate approach is better suited to describe the price difference distribution than a univariate approach. 
In the following, the term historical selection refers to the multivariate selection of historical data.    \section{Multivariate probabilistic forecasting of electricity price differences}
\label{sec:Price_Delta_DistributionModeling}

The realizations of the hourly price difference vector in Equation~\eqref{Eq:Price_Delta_Definition_Price_Difference_vector} follow an unknown four-dimensional joint distribution:
\begin{equation} \label{Eq:Price_Delta_Definition_Price_Difference_PDF}
    \bm{\Delta}\textbf{ID}_3 \sim p_{\bm{\Delta}\textbf{ID}_3} \left(\bm{\Delta}\textbf{ID}_3 \right)
\end{equation}
Here, $p_{\bm{\Delta}\textbf{ID}_3} \left(\bm{\Delta}\textbf{ID}_3 \right)$ is the probability density function (PDF) of the vector-valued random variable $\bm{\Delta}\textbf{ID}_3$.
The realizations of the price difference distribution are assumed to be influenced by known external factors.
Therefore, the proposed modeling approach includes these external factors as input features $\mathbf{y}$ to form a conditional distribution:
\begin{equation} \label{Eq:Price_Delta_Definition_Price_Difference_PDF_Exogenious}
    \bm{\Delta}\textbf{ID}_3 \sim p_{\bm{\Delta}\textbf{ID}_3 \vert Y} \left(\bm{\Delta}\textbf{ID}_3\vert \mathbf{y} \right)
\end{equation}
Here, $p_{\bm{\Delta}\textbf{ID}_3 \vert Y} \left(\bm{\Delta}\textbf{ID}_3\vert \mathbf{y} \right)$ is the conditional PDF of $\bm{\Delta}\textbf{ID}_3$ given the input feature $\mathbf{y}$.
Forecasting this conditional multivariate probability distribution poses a difficult modeling problem. The probability distribution is of unknown shape and electricity prices are known to exhibit heavy tails \citep{han2022complexity}. Furthermore, the relationship with the conditional input factors is unknown and likely nonlinear. Thus, the proposed probabilistic forecasting approach requires a flexible distribution model that can seamlessly include external input features with potentially nonlinear effects. 

To model the conditional probability distribution of price differences described by Equation~\eqref{Eq:Price_Delta_Definition_Price_Difference_PDF_Exogenious}, this work uses the following four approaches: (1) conditional normalizing flows \citep{winkler2019learning}, (2) the informed selection of historical data presented in Section~\ref{sec:Price_Delta_Methodology_4D_modeling}, (3) Gaussian copulas with quantile regression \citep{pinson2009probabilistic}, and (4) multivariate Gaussian regression \citep{dillon2017tensorflow}. 
The selection of models aims to provide a variety of na\"ive, established, and recently published approaches.

\begin{figure}
    \centering
    \includegraphics[width=\columnwidth]{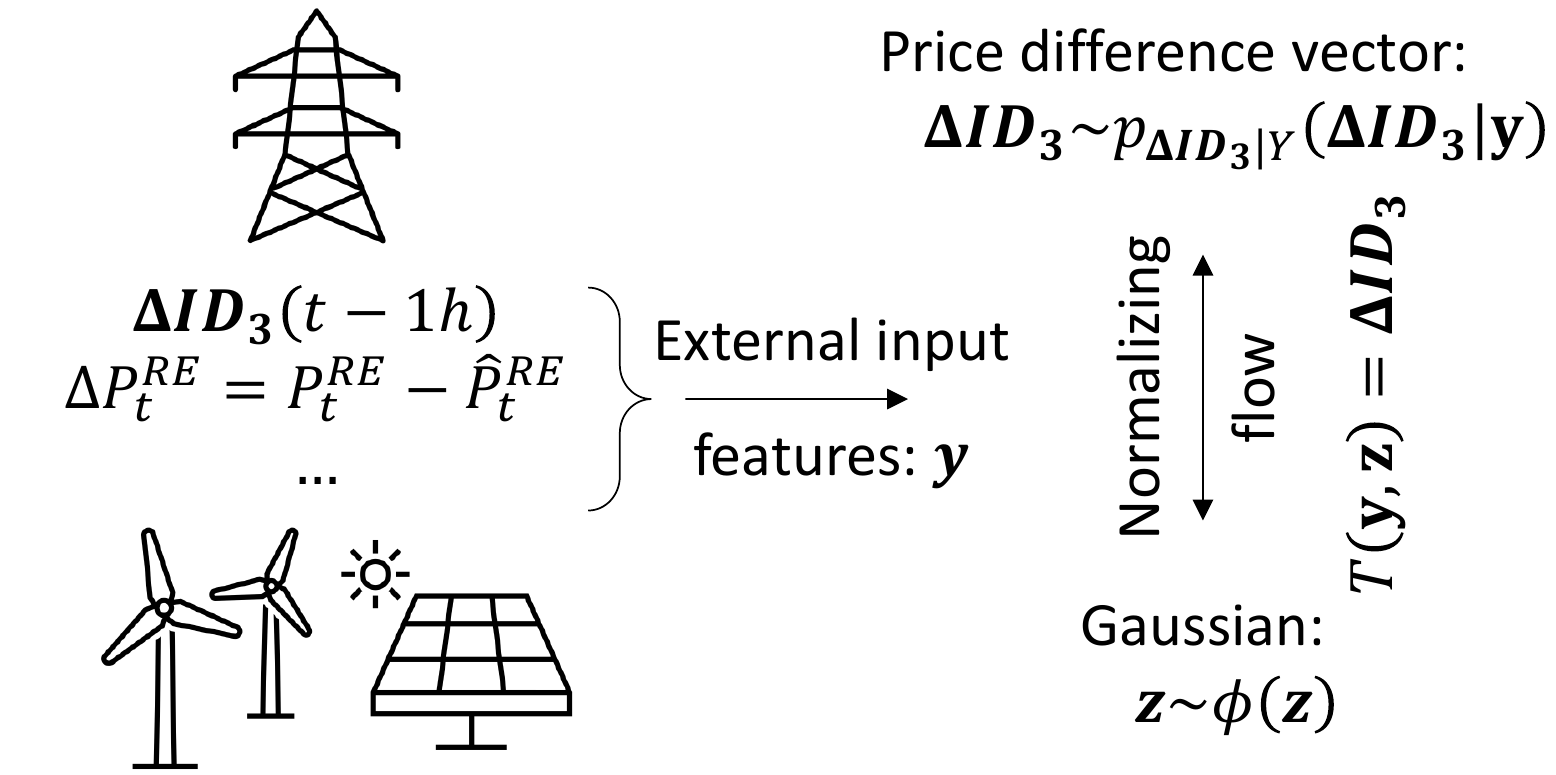}
    \caption{Multivariate probabilistic forecasting scheme. The external input features are collected in a feature vector $\mathbf{y}$ and used to condition the transformation of Gaussian samples via the normalizing flow.}
    \label{fig:Price_Delta_NF_Scheme}
\end{figure}

Normalizing flows are non-parametric distribution models that model high-dimensional complex distributions as an invertible neural network transformation $T$, with inverse $T^{-1}$, of a multivariate standard Gaussian $\phi(\mathbf{z})$ \citep{dinh2016realNVP, papamakarios2021normalizing}. Utilizing the change of variables formula, normalizing flows describe the conditional PDF $p_{X\vert Y}(\mathbf{x} \vert \mathbf{y})$ of a multivariate random variable $X$ with realizations $\mathbf{x}$ and input features $\mathbf{y}$ explicitly \citep{papamakarios2021normalizing, winkler2019learning}:
\begin{equation}
    p_{X\vert Y}(\mathbf{x} \vert \mathbf{y}) = \phi\left(T^{-1}(\mathbf{x}, \mathbf{y})\right) \left|\det \mathbf{J}_{T^{-1}}(\mathbf{x}, \mathbf{y}) \right|
\end{equation}
Here, $\mathbf{J}_{T^{-1}}$ is the Jacobian of the inverse transformation. 
Normalizing flows make no assumptions about the data structure and distribution. Thus, by using sufficiently expressive transformations, normalizing flows can model any distribution \citep{papamakarios2021normalizing}.
To sample from a normalizing flow, first, samples are drawn from the multivariate Gaussian, i.e., $\hat{\mathbf{z}}_i \sim \phi(\mathbf{z})$. Second, the samples and the conditional inputs are transformed using the forward form of the invertible neural network:
\begin{equation*}
    \hat{\mathbf{x}}_i = T\left(\hat{\mathbf{z}}_i, \hat{\mathbf{y}} \right) \quad \forall i=1,2,\dots,N
\end{equation*}
Here, $\hat{\mathbf{y}}$ are the external input features at the respective time, which are assumed to be given. 
Figure \ref{fig:Price_Delta_NF_Scheme} shows a sketch of the conditional normalizing flow modeling scheme. 
This work implements the invertible neural network using the real non-volume preserving transformation (RealNVP) \citep{dinh2016realNVP}.

The Gaussian copula is a well-established method to describe multivariate distributions with nontrivial correlations between the individual dimensions. The basic approach relies on transforming samples from a uniform distribution via an estimated inverse cumulative distribution function (CDF). In forecasting applications, the inverse CDF is estimated via an interpolation of quantile forecasts from quantile regression. For a detailed introduction to the combination with quantile regression, we refer the interested reader to \cite{pinson2009probabilistic}. The present work uses the quantile regression implemented in the `statsmodels' library \citep{seabold2010statsmodels}.

Finally, the multivariate Gaussian regression uses neural networks to predict the mean vector $\bm{\mu}_X$ and the covariance matrix $\bm{\Sigma}_X$ of the multivariate Gaussian as a function of the conditional inputs $\mathbf{y}$ \citep{dillon2017tensorflow}:
\begin{equation}
    p_{X\vert Y}(\mathbf{x} \vert \mathbf{y}) = \mathcal{N}_{X\vert Y}\left(\mathbf{x}; \bm{\mu}_X(\mathbf{y}), \bm{\Sigma}_X(\mathbf{y}) \right)
\end{equation}
As the covariance matrix is symmetric, the neural network only needs to predict a lower-triangular version of $\bm{\Sigma}_X(\mathbf{y})$ \citep{dillon2017tensorflow}.
The predicted mean and covariance can then be used to sample from the Gaussian distribution parameterized by the predicted values.

In all models, the electricity price difference vector $\bm{\Delta}\textbf{ID}_3$ takes the role of the random variable $X$, and the external factors are the conditional input features $Y$.
For every realization of the external factors $\mathbf{y}$, the models allow sampling from the electricity price difference vector distribution $\hat{\bm{\Delta}\textbf{ID}_3}\sim p_{\bm{\Delta}\textbf{ID}_3}$ as described for the normalizing flow above.    \section{Input feature selection}\label{sec:Price_Delta_InputFeatureSelection}
The realizations of both the day-ahead and the intraday prices are influenced by external factors such as the residual loads \citep{kiesel2017econometric}. 
However, the driving factors for the electricity price difference vector are likely to be different compared to those for the absolute day-ahead and intraday prices. 
In particular, the difference alleviates the impact of some external factors and amplifies the impact of others. 
Hence, the selection of input features should not be based on the standard inputs for day-ahead or intraday price forecasting known in the literature.

This Section aims to find a highly informative set of external factors that should be included as input features to the models in Section~\ref{sec:Price_Delta_DistributionModeling}. 
First, Section~\ref{sec:Price_Delta_DerivingFeatureVector} derives a set of possible input feature based on literature impressions and an analysis of the price time series. 
Second, Section~\ref{sec:Delta_Price_XAI} uses explainable artificial intelligence (XAI) to narrow down the full set to a highly informative subset of input features.

\subsection{Possible input features} \label{sec:Price_Delta_DerivingFeatureVector}
The day-ahead prices are determined by the intersection of supply and demand \citep{pape2016fundamentals,gurtler2018effect} and influenced by the merit-order curve \citep{kremer2021econometric}. 
The intraday markets are used to adjust the previously submitted bids to address short-term changes in supply and demand. 
Thus, the impact of the forecasted total electricity demand and the forecasted total renewable electricity supply is already reflected in the day-ahead prices and has little influence on the difference between day-ahead prices and intraday prices \citep{kiesel2017econometric, spodniak2021impact, han2022complexity}.
The differences are mostly impacted by the forecasting errors in the residual load, i.e., for load and renewable electricity generation \citep{kiesel2017econometric,han2022complexity}. For the absolute intraday prices, Kalukov and Ziel \citep{kulakov2019impact} show that the impact of forecast errors is nonlinear. 
Notably, the actual renewable forecast errors are only known a posteriori. Hence, these values are neither known to intraday traders, nor can they be applied in actual forecasts of the intraday prices. However, hour-ahead renewable forecast are available \citep{sweeney2020future} and can be used to correct day-ahead forecasts, thus affecting the intraday electricity prices. Unfortunately, these hour-ahead renewable forecasts are not publicly available. Hence, we use the a posteriori renewable forecasting errors as a proxi in our analysis:
\begin{equation*}
    \Delta P^{RE}_t = P^{RE}_t - \hat{P}^{RE}_t\quad RE \in \{\text{Solar}, \text{Wind}, \text{Load}\} 
\end{equation*}
Here, $P^{RE}_t$ is the actual, $\hat{P}^{RE}_t$ is the day-ahead forecast.

Another important external factor for the fluctuation of the intraday prices is the ramping of renewables and load \citep{koch2019short}. 
For instance, solar electricity generation ramps up in the morning and ramps down in the evening, which leads to large changes in solar power supply within single day-ahead trading intervals \citep{koch2019short}. 
\cite{wolff2017short} and \cite{ocker2017german} find that the ratio of conventional and renewable electricity impacts the realizations of both day-ahead and intraday prices. To investigate the impact of the share of conventional electricity, the prices of oil and natural gas are considered as input factors. 

\cite{han2022complexity} find that the increments of the day-ahead price, i.e., the difference between one hour and the next, have a substantial impact on the direction (increasing or decreasing) and the intensity of the hourly fluctuation. 
This observation is confirmed by the average daily price profiles shown in Figure~\ref{fig:Price_Delta_AverageDay}. For increasing or decreasing day-ahead prices, the hourly fluctuation shows increasing or decreasing patterns, respectively. Furthermore, larger increments appear to induce more significant fluctuations and vice versa. 
For local minimum and maximum peak hours in the day-ahead price, Figure~\ref{fig:Price_Delta_AverageDay} shows no distinct trend of the fluctuation. 

Figure~\ref{fig:Price_Delta_AverageDay} also indicates that the realizations of the price differences depend on the time of the day. 
Thus, the list of possible input features includes a trigonometric encoding of the hour of the day \citep{moon2019comparative}.

\begin{figure}
    \centering
    \includegraphics[width=\columnwidth]{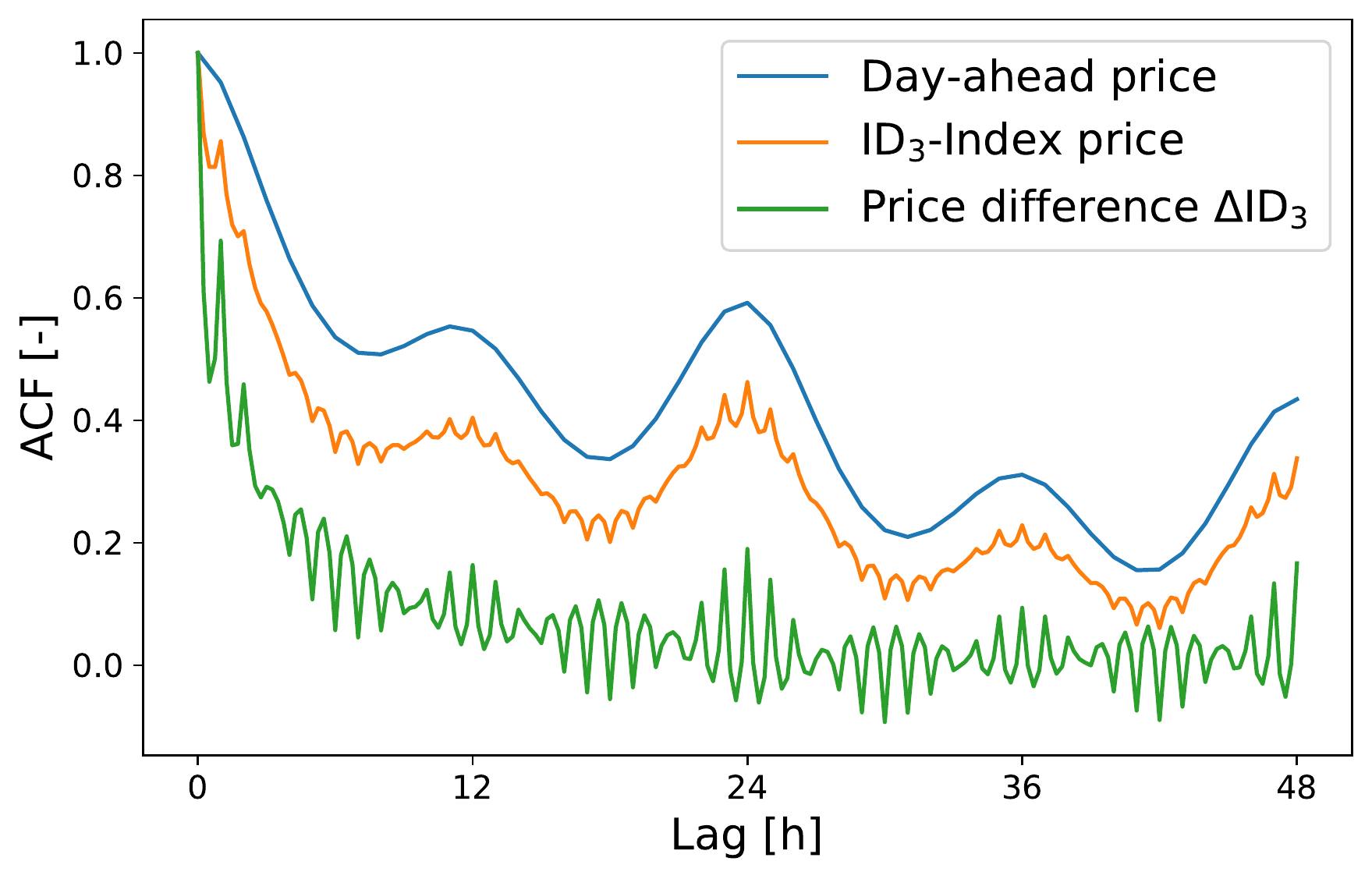}
    \caption{Pearson autocorrelation (ACF) of day-ahead price (``Day-ahead price''), ID$_3$ index (``ID$_3$-Index price''), and difference between the two time series (``Price difference $\Delta$ID$_3$'') with time lag of 0 to 48\,h. 
   EPEX spot price data in 2018 and 2019 from \cite{EnergyCharts2022}.
    }
    \label{fig:Price_Delta_ACF}
\end{figure}
\cite{narajewski2020econometric} and \cite{han2022complexity} state that the day-ahead markets in Germany are \emph{weak-form efficient}, i.e., the recent history of day-ahead prices does not inform on future realizations, whereas for intraday prices, there is an advantage from including the most recent value in the prediction. 
To evaluate the impact of prior realizations on the price difference, Figure~\ref{fig:Price_Delta_ACF} shows the Pearson autocorrelation function (ACF) of the day-ahead price, the intraday price, and the difference between the two time series for time lags up to 48\,h. 
The ACF of the price difference drops quickly to values that are significantly lower compared to those for the day-ahead ACF and the intraday ACF that persist at high levels over 0.3 over longer time spans. 
After the rapid drop-off, the price difference ACF fluctuates at a low level under 0.2 indicating negligible impact for forecasting tasks. 
Yet, the first few hours still show high ACF values. Thus, the two previous hours are included as possible features. 
The supplementary material of this work shows further analysis of the time dependency on longer scales. 
Notably, the price difference time series does not show any significant trends over the days of the week or the months of the year over the considered dataset.

In the following, the possible input features are considered in three groups: The forecast error features, the ramping features, and the price and time features. The considered feature labels and their descriptions are listed in Table~\ref{tab:Price_Delta_Feature_Labels_PriceTime}, \ref{tab:Price_Delta_Feature_Labels_Errors}, and \ref{tab:Price_Delta_Feature_Labels_Ramping}, respectively.

\begin{table}
\centering
\caption{Price and time feature labels. The variable $t$ denotes the considered time interval in hours. }
\label{tab:Price_Delta_Feature_Labels_PriceTime}
\resizebox{\columnwidth}{!}{\begin{tabular}{ll}
\toprule
Description                      & Labels                         \\ \midrule
Day Ahead Auction                & $DA$                      \\
Day Ahead Increments back        & $\Delta DA^{-}$           \\
Day Ahead Increments ahead       & $\Delta DA^{+}$           \\
cosine time encoding             & cos(t)                    \\
sine time encoding               & sin(t)                    \\
$DA - ID_3$ at t-1\,h (1$^{\text{st}}$ interval) & $\Delta ID_3^{00} (t-1h)$ \\
$DA - ID_3$ at t-1\,h (2$^{\text{nd}}$ interval) & $\Delta ID_3^{15} (t-1h)$ \\
$DA - ID_3$ at t-1\,h (3$^{\text{rd}}$ interval) & $\Delta ID_3^{30} (t-1h)$ \\
$DA - ID_3$ at t-1\,h (4$^{\text{th}}$ interval) & $\Delta ID_3^{45} (t-1h)$ \\
$DA - ID_3$ at t-2\,h (1$^{\text{st}}$ interval) & $\Delta ID_3^{00} (t-2h)$ \\
$DA - ID_3$ at t-2\,h (2$^{\text{nd}}$ interval) & $\Delta ID_3^{15} (t-2h)$ \\
$DA - ID_3$ at t-2\,h (3$^{\text{rd}}$ interval) & $\Delta ID_3^{30} (t-2h)$ \\
$DA - ID_3$ at t-2\,h (4$^{\text{th}}$ interval) & $\Delta ID_3^{45} (t-2h)$ \\
\bottomrule
\end{tabular} }
\end{table}

\begin{table}
\centering
\caption{Forecast error feature labels.}
\label{tab:Price_Delta_Feature_Labels_Errors}
\resizebox{\columnwidth}{!}{\begin{tabular}{ll}
\toprule
     Description    & Labels                         \\ \midrule
Solar Error {[}MW{]} (1$^{\text{st}}$ interval)           & Solar Error 00            \\
Solar Error {[}MW{]} (2$^{\text{nd}}$ interval)           & Solar Error 15            \\
Solar Error {[}MW{]} (3$^{\text{rd}}$ interval)           & Solar Error 30            \\
Solar Error {[}MW{]} (4$^{\text{th}}$ interval)           & Solar Error 45            \\
Load Error {[}MW{]} (1$^{\text{st}}$ interval)            & Load Error 00             \\
Load Error {[}MW{]} (2$^{\text{nd}}$ interval)            & Load Error 15             \\
Load Error {[}MW{]} (3$^{\text{rd}}$ interval)            & Load Error 30             \\
Load Error {[}MW{]} (4$^{\text{th}}$ interval)            & Load Error 45             \\
Wind Error {[}MW{]} (1$^{\text{st}}$ interval)   & Wind Error 00             \\
Wind Error {[}MW{]} (2$^{\text{nd}}$ interval)   & Wind Error 15             \\
Wind Error {[}MW{]} (3$^{\text{rd}}$ interval)   & Wind Error 30             \\
Wind Error {[}MW{]} (4$^{\text{th}}$ interval)   & Wind Error 45             \\ 
\bottomrule
\end{tabular}}
\end{table}

\begin{table}
\centering
\caption{Day-ahead ramping feature labels.}
\label{tab:Price_Delta_Feature_Labels_Ramping}
\resizebox{\columnwidth}{!}{\begin{tabular}{ll}
\toprule
     Description    & Labels                         \\ \midrule
Solar  & $DA$ solar ramp           \\
Wind & $DA$ wind ramp            \\
Load   & $DA$ load ramp            \\
Total power generation  & $DA$ total gen ramp       \\
Import/export    & $DA$ import/export ramp   \\
\bottomrule
\end{tabular}
}
\end{table}

\subsection{Input feature selection via explainable AI}\label{sec:Delta_Price_XAI}
To improve usability and to avoid overfitting of the forecasting models, the set of input features should be narrowed down to a highly informative subset. 
While relevant external impact factors for day-ahead and intraday prices are well studied within the literature \citep{wolff2017short, trebbien2022understanding}, there is no evaluation of impact factors for the difference between the two prices. 
The previous Section lists a number of possible input features. However, the true effect on the realization of the price differences remains unknown.
This work uses XAI to find the most impactful input features. In particular, the explanatory power of SHAP-values is used to quantify the impact of each input feature and, thus, make informed decisions for the feature set selection.

Given a model $f(\mathbf{y})$ with feature vector $\mathbf{y}$, the SHAP-value approach \citep{lundberg2020local} decomposes each model prediction into the contributions of the individual features:
\begin{equation}
    f(\mathbf{y}) =  \varphi_0 + \sum_i  \varphi_i(y_i;f),
\end{equation}
where $\varphi_0 = \mathbb{E}_Y(f(\mathbf{y}))$ is a constant offset and $\varphi_i$ are the SHAP-values of the respective input features $y_i\in \mathbf{y}$. The SHAP-value decomposition ensures consistent interpretability, and the individual SHAP-values quantify the contribution of each input feature towards the model output value. 
The SHAP value method is designed specifically to quantify the impact of each input feature considering all possible (nonlinear) combinatorial effects represented in the model $f(\mathbf{y})$. 
Hence, SHAP values are not restricted to simple marginal feature perturbations under the ceteris parbus assumption. They capture the contribution of a given feature for all possible coalitions of other features in a consistent way, such that there is no possibility for missing combinatorial effects of different features changing the model outputs.Thus, the SHAP value method evaluates how each feature impacts the effects of the other features on the model output. 
In the supplementary material, we provide more information on the SHAP-value method. For a detailed introduction, the reader is referred to the original work by \cite{lundberg2020local}.

SHAP-values provide local information for each element of the data set. A global measure of the importance of each feature can be obtained by taking the average of the absolute values:
\begin{equation}\label{Eq:Price_Delta_Feature_Importance}
    \text{FI}_i = \sum_{t \in \chi_\text{test}} \vert \varphi_i(y(t);f) \vert 
\end{equation}
Here, $\text{FI}_i$ is the feature importance of the $i$-th input feature, $\vert\cdot\vert$ denotes the absolute value operator, and $\chi_\text{test}$ is the test set.
Notably, SHAP-values do not inform on the quality of the prediction itself.
For a detailed introduction to SHAP-values and XAI, the reader is referred to \cite{lundberg2020local} and to \cite{kruse2021revealing} for an example application to power grid frequencies.

The SHAP value concept is designed for deterministic regression models. Thus, SHAP-values cannot be computed for the probabilistic models in Section~\ref{sec:Price_Delta_DistributionModeling}. 
Instead, this work uses a tree regressor model, which is a popular choice to compute SHAP-values \citep{lundberg2020local}. In particular, gradient boosted trees have been shown to offer high levels of interpretability \citep{lundberg2020local,kruse2021revealing}. 
The gradient boosted trees are trained on 80\% of the days from 2018 and 2019 and the remaining 20\% of days are set aside as a test set. The test set is then used to score the accuracy of the trained models and to compute the SHAP-values. 
As SHAP-values can only be computed for scalar regressors, each of the four dimensions of the price difference vector is considered separately. 
For the SHAP-value analysis, this work uses the `SHAP' library \citep{shap_library} and the gradient boosted tree implementation in the machine-learning library scikit-learn \citep{scikitlearn}.
For the test set, the regressor returns $R^2$ of 74\%, 60\%, 61\%, and 63\% for each of the four price differences, respectively.

\begin{figure}
    \centering
    \includegraphics[width=\columnwidth]{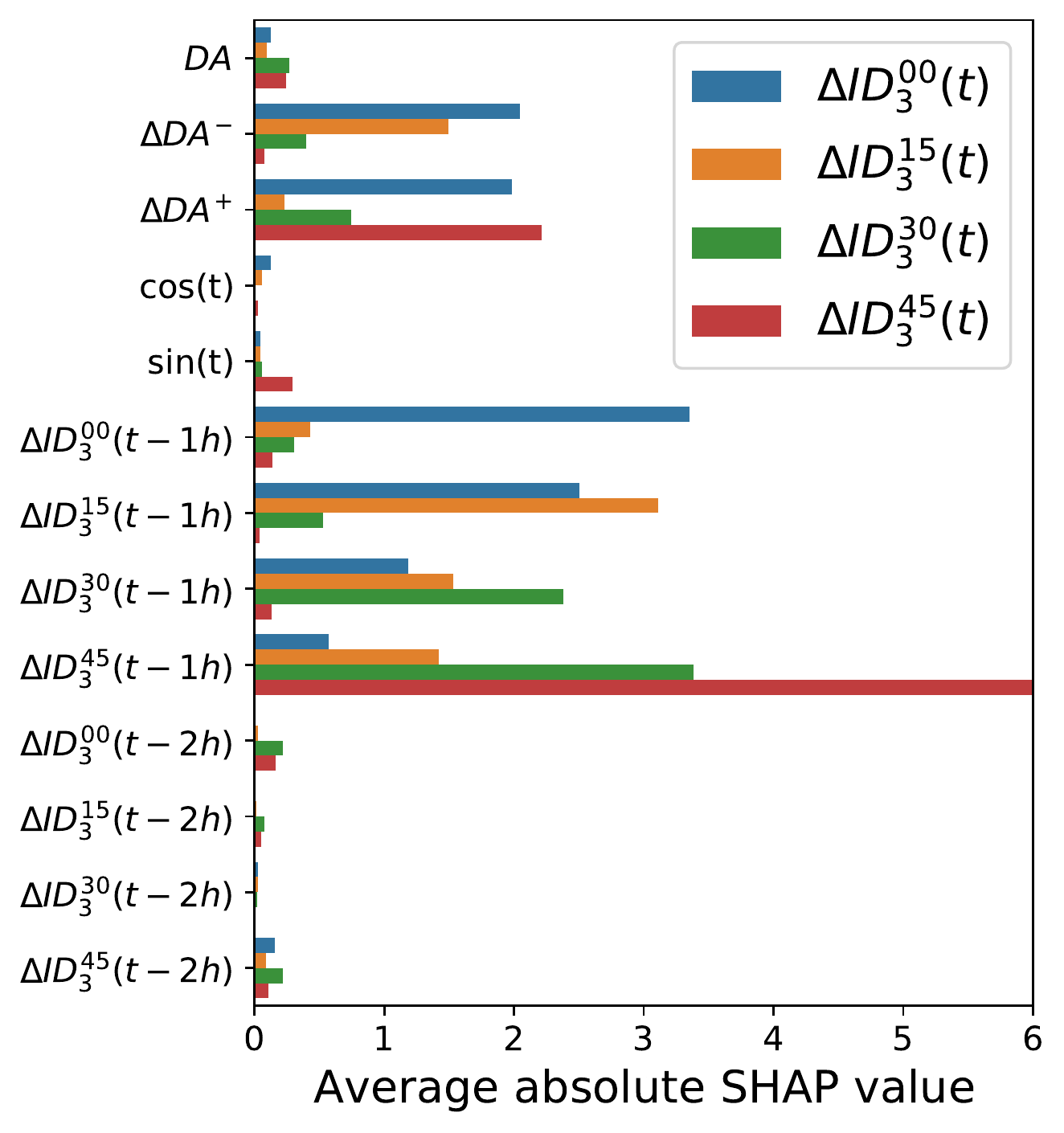}
    \caption{
        SHAP-values \citep{lundberg2020local} of the price difference vector for the price and time features. $DA$: day-ahead price, $\Delta DA^{+/-}$: day-ahead increments ahead and back, sin(t)/cos(t): trigonemetric hour encodings, and $\Delta ID_3^{XX}$: price difference at given hour and 15\,min interval.
    }
    \label{fig:Price_Delta_SHAP_Price}
\end{figure}

\begin{figure}
    \centering
    \includegraphics[width=\columnwidth]{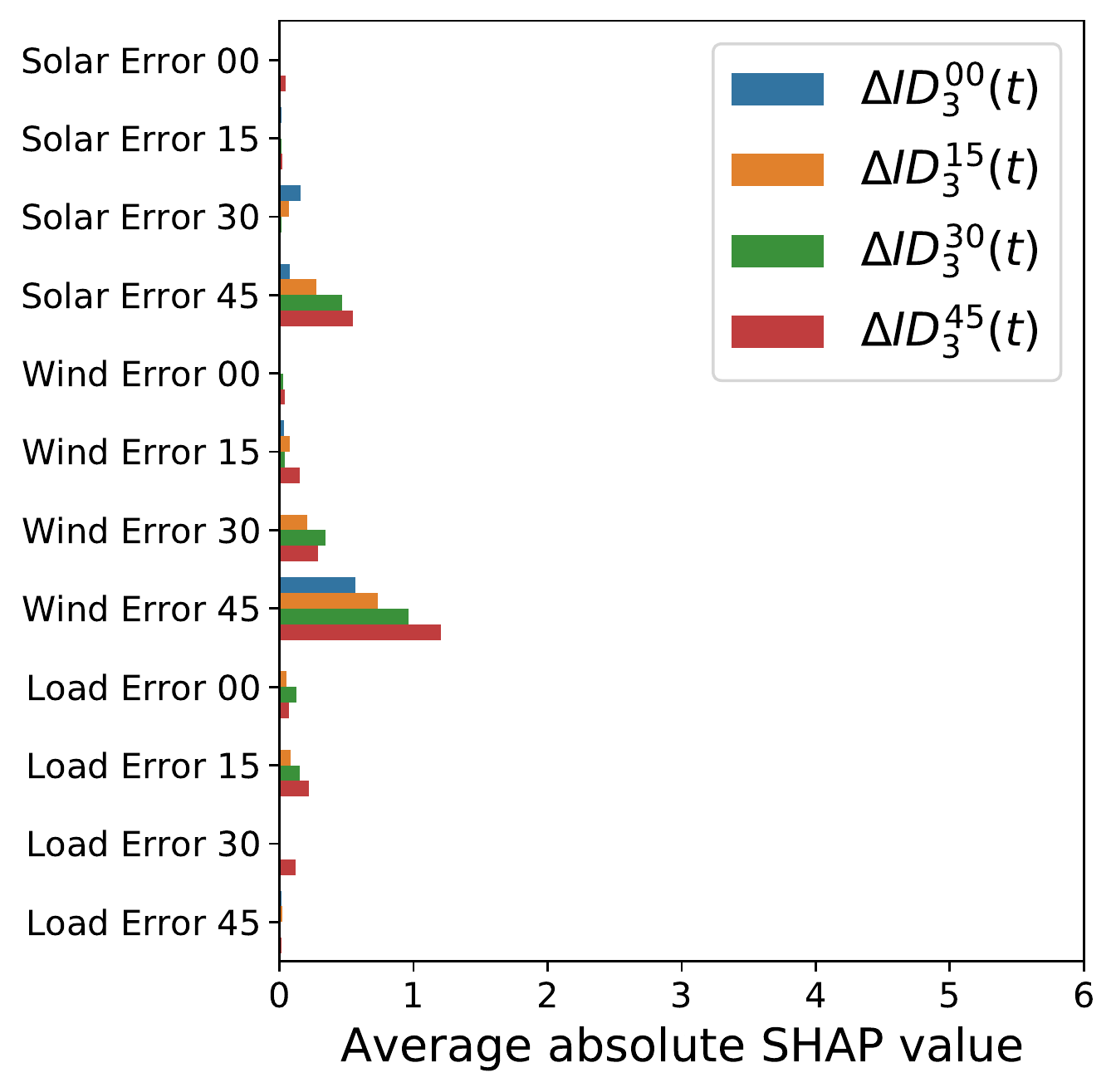}
    \caption{
        SHAP-values \citep{lundberg2020local} of the price difference vector for the forecast error features. Solar, Wind, and Load: forecast errors at given hour and 15\,min interval.
    }
    \label{fig:Price_Delta_SHAP_Error}
\end{figure}

\begin{figure}
    \centering
    \includegraphics[width=\columnwidth]{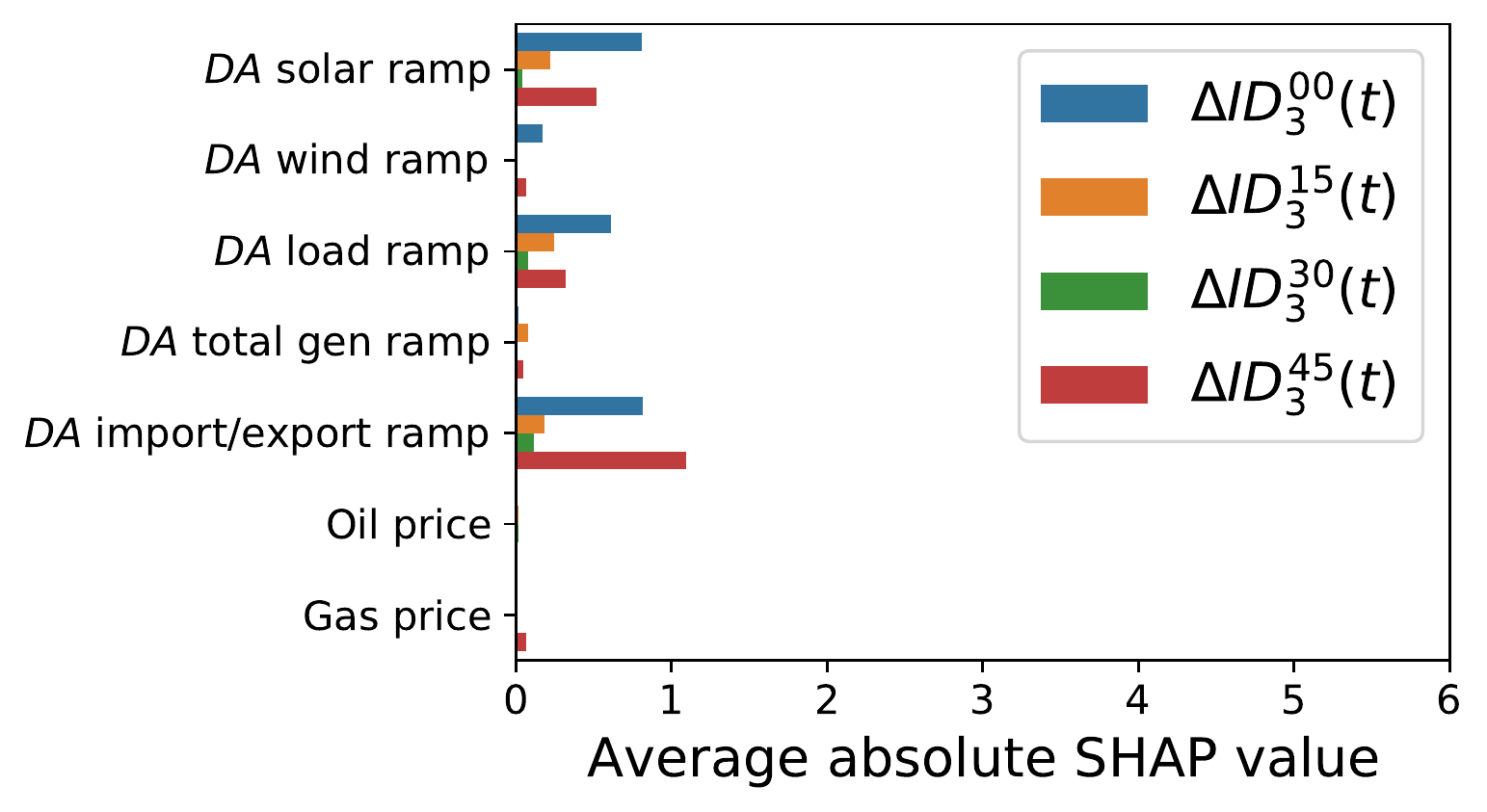}
    \caption{
        SHAP-values \citep{lundberg2020local} of the price difference vector for the day-ahead ramping features. Day-ahead ramps for solar, wind, load, total power generation, and import and export balances. 
    }
    \label{fig:Price_Delta_SHAP_Ramps}
\end{figure}

The SHAP-values are computed using a single regressor for all input features.
Figure~\ref{fig:Price_Delta_SHAP_Price}, Figure~\ref{fig:Price_Delta_SHAP_Error}, and Figure~\ref{fig:Price_Delta_SHAP_Ramps} show the feature importance defined in Equation~\eqref{Eq:Price_Delta_Feature_Importance} for price and time features, the forecast error features, and the ramping features, respectively. 
The SHAP-values imply that there are few highly impactful features and many negligible features. In particular, the realizations of the price difference vector in the previous hour and the day-ahead increments show the most significant impact. 
The previous hour shows the highest SHAP-values for the same 15\,min interval, respectively.
Meanwhile, the absolute day-ahead price, the trigonometric time encoding, the fossil fuel prices, and the second to last price difference vector realizations also show negligible impact.
Of the forecast errors, only the renewables forecast errors in the last 15 minute interval show considerable impact. Note that using hour-ahead forecasts may change the feature importance for the forecast errors. However, the small difference between the hour-ahead forecasts and the realizations combined with the low feature importance of the proxy indicate that considering the hour-ahead forecast is unlikely to cause significant changes. 
The ramping features mostly impact the first and the fourth interval in the price difference vector. Thus, the ramps mostly impact the orientation of the increasing or decreasing pattern in the data. 
Notably, the overall impact of the ramping features is minor.

In summary, the SHAP-value analysis indicates that the day-ahead increments and the realization of the previous hour are the most informative as input features to the models in Section~\ref{sec:Price_Delta_DistributionModeling}.
The analysis further suggests that the performance of models trained to predict the price difference is independent of most external impact factors. For instance, oil and gas prices have negligible feature importance for the price differences.
Intraday trading decisions are made in situations of immediate requirement rather than a planning situation. Thus, the differences to the day-ahead prices are not affected by fuel prices. 
Still, electricity markets are continuously evolving, and the SHAP results in this study do not necessarily translate to other times. 

For an analysis of the Pearson correlation of the impact features, see the supplementary material of this work.    \section{Numerical experiments}\label{sec:Price_Delta_Results}
This Section applies the different probabilistic modeling approaches to predict the price difference vector distribution. 
Each model is trained on the data set from January 2018 to June 2019 \citep{EnergyCharts2022}, and the month of July 2019 is used as a test set. The 24\,h $\times$ 31-day test set results in 744 test intervals.
Note that July 1st, 2019, was a Monday. 
To preprocess the data and avoid complications from the heavy-tailed distributions \citep{han2022complexity}, the data is transformed using the probability integral transform discussed in \cite{uniejewski2017variance}, which renders the data in a Gaussian form. 
The normalizing flow and the Gaussian regression are trained via log-likelihood maximization over 500 epochs with a batch size of 128 using the python-based machine-learning library TensorFlow version 2.8.0 \citep{tensorflow2015}.
The Gaussian copula uses the quantile regression in the `statsmodels' python library \citep{seabold2010statsmodels}.
For details on the implementation of the models, see the supplementary material. 
One hundred samples are drawn for each hour in the test set and used to compute the different metrics outlined in Section~\ref{sec:Price_Delta_Methodology_4D_modeling}. 
Sections~\ref{sec:Price_Delta_Prediction_Intervals} and \ref{sec:Price_Delta_Energy_Variogramm} use the full set of input features derived in Section~\ref{sec:Price_Delta_DerivingFeatureVector}. Section~\ref{sec:Delta_Price_Empirical_Significants_Features} performs an empirical investigation of the impact of the different groups of input features.

\subsection{Prediction intervals}\label{sec:Price_Delta_Prediction_Intervals}
This Section analyzes the forecasted distributions by investigating the 50\% and 90\% prediction intervals. 
The intervals are estimated by computing the respective quantiles of the drawn samples for each time step and then adding the day-ahead price of the given hour to reverse the difference in Equation~\eqref{Eq:Price_Delta_Definition_Price_Difference_vector}. 
\begin{figure*}
    \centering
    \includegraphics[width=0.93\textwidth]{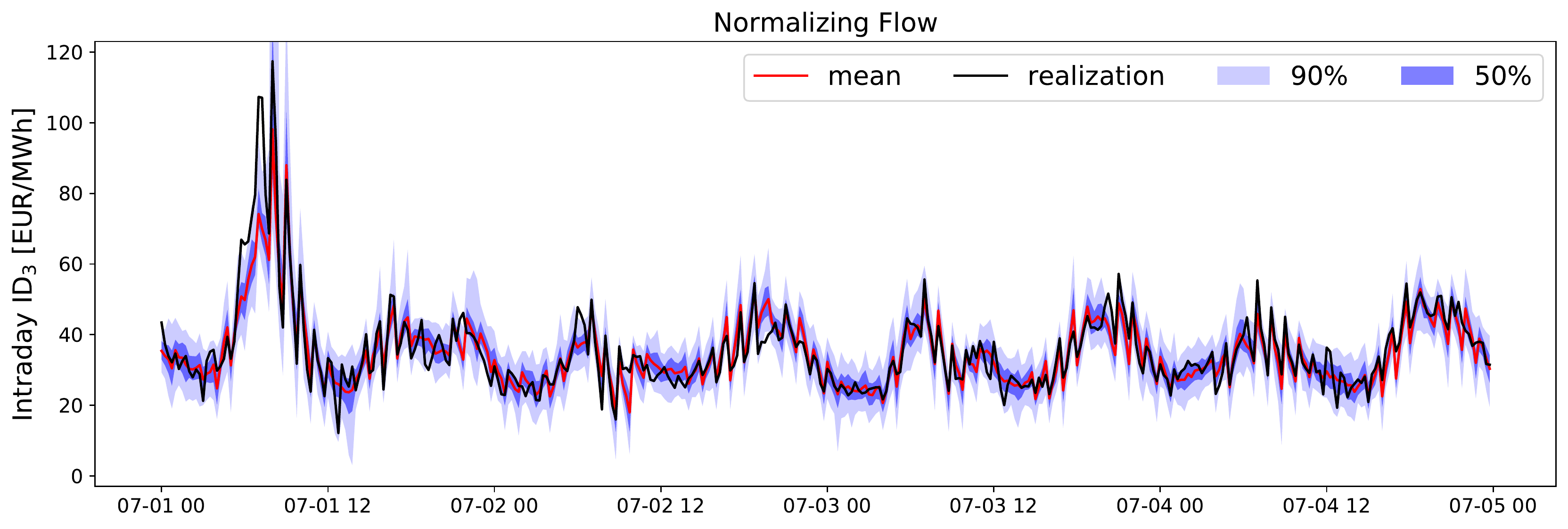}
    \includegraphics[width=0.93\textwidth]{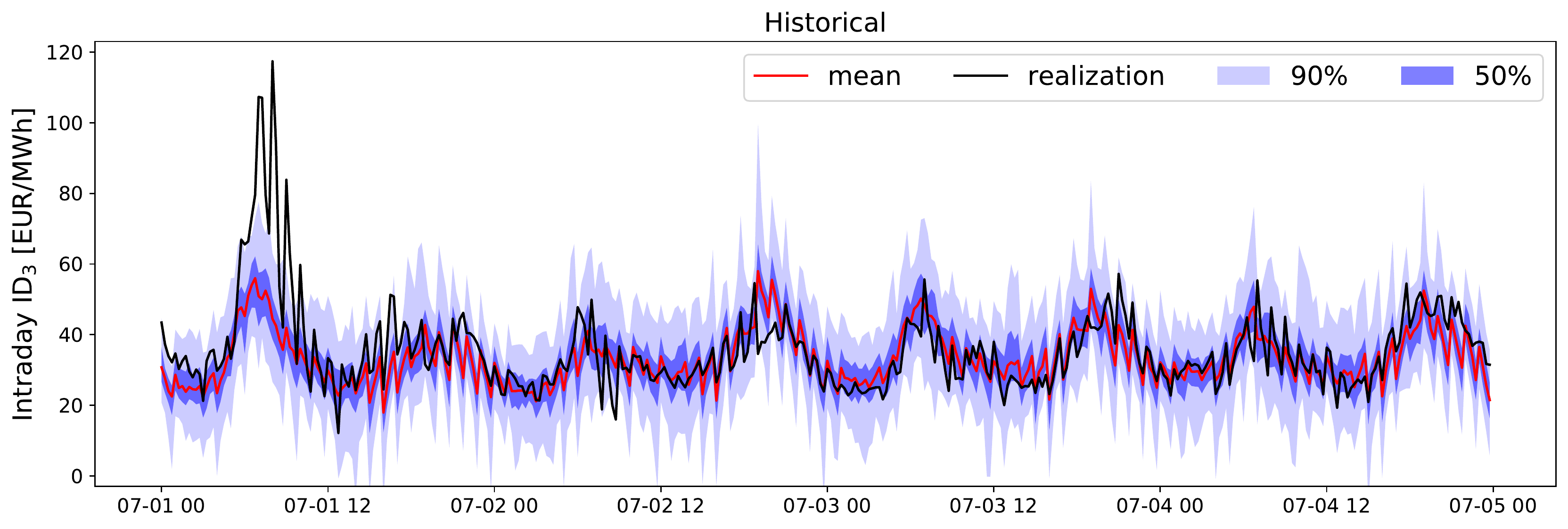}
    \includegraphics[width=0.93\textwidth]{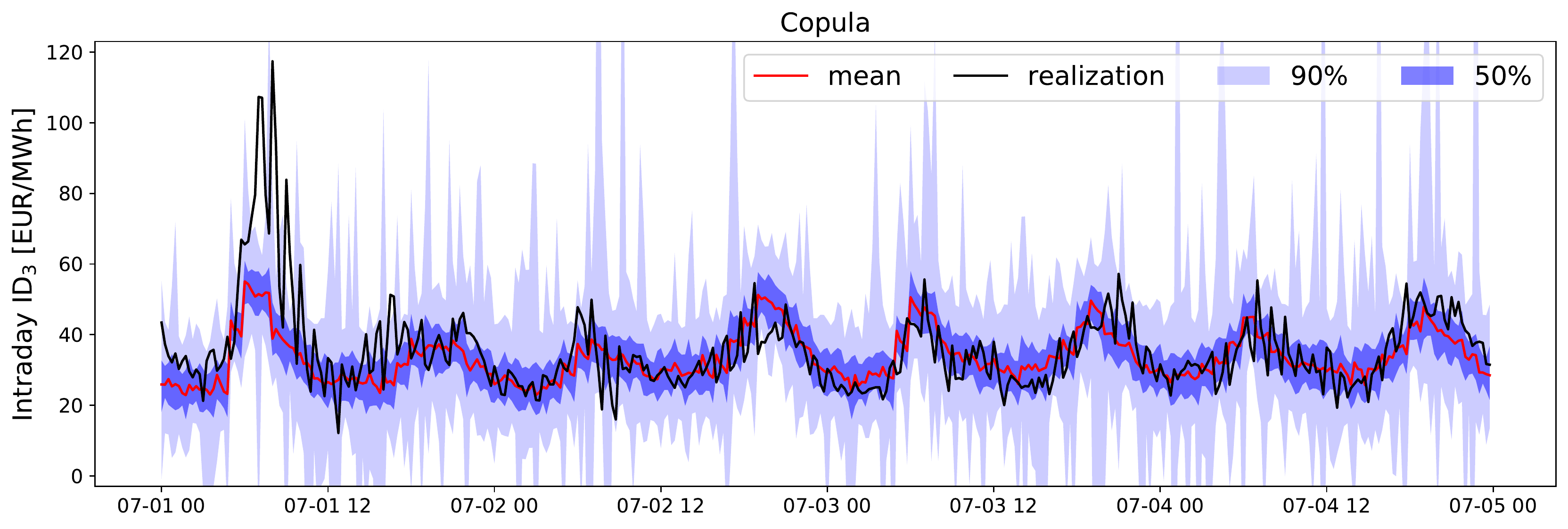}
    \includegraphics[width=0.93\textwidth]{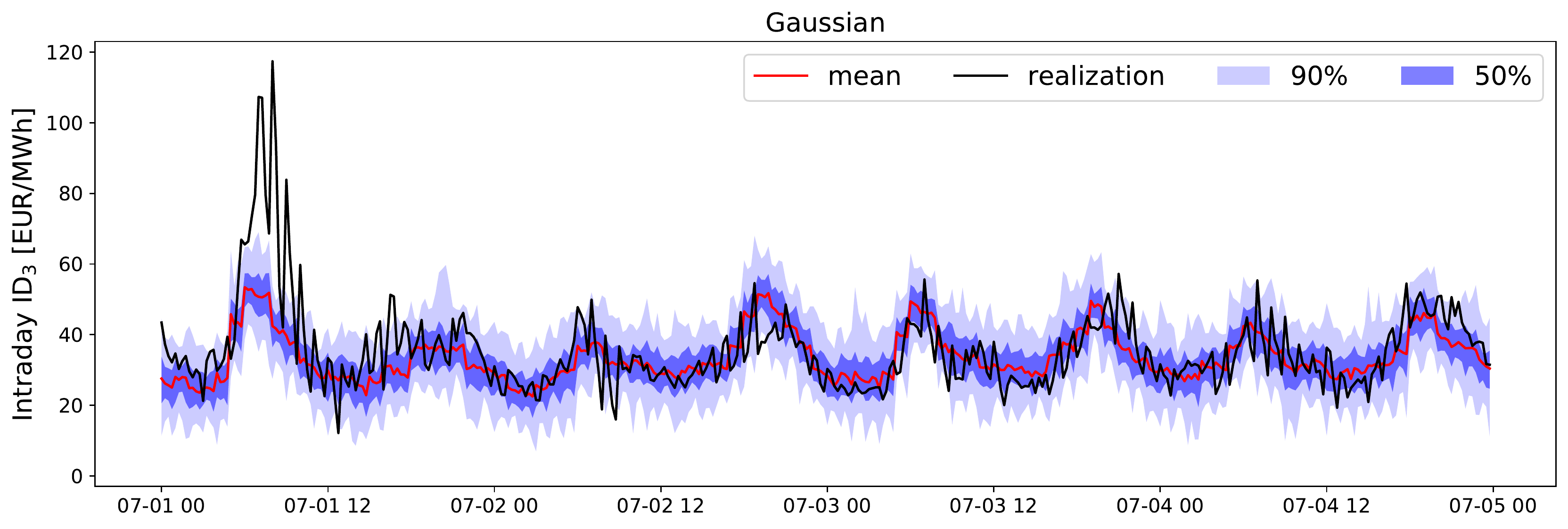}
    \caption{
Comparison of probabilistic forecasts by the normalizing flow, the informed selection of historical data, The Gaussian copula, and the Gaussian regression model to the actual realization (black line) for the first five days in July 2019. The probabilistic forecasts are visualized in terms of the median (red line) and the 90\% and 50\% quantiles (shaded areas), respectively.
    Note the truncated y-axis for the Gaussian copula.
    Training data from January 2018 to June 2019 \citep{EnergyCharts2022}.
    Results generated with all conditional inputs derived in Section~\ref{sec:Price_Delta_DerivingFeatureVector}.
    }
    \label{fig:Price_Delta_Example_Prediction_Intervals}
\end{figure*}

Figure~\ref{fig:Price_Delta_Example_Prediction_Intervals} shows the predicted mean, the 50\%, and 90\% prediction intervals, and the intraday price realization for the first four days in the test month of July 2019. The Figure shows the results estimated from probabilistic forecasts from the normalizing flow (top), the selection of historical data (second from top), the Gaussian copula (third from top), and the Gaussian regression (bottom).
For the presented test days, the normalizing flow prediction intervals enclose the realizations for most of the time steps, and the forecasted mean identifies the realized trends well, with very few exceptions. 
The selection of historical data reflects the trends moderately well and portrays wider prediction intervals compared to the normalizing flow. 
The Gaussian copula and the Gaussian regression often fail to identify the correct trends and do not fit the realizations as tightly as the normalizing flow. In particular, the Gaussian copula results in extensive prediction intervals that show vast peaks for the 90\% prediction interval. 
Compared to the other approaches, the normalizing flow performs significantly better in identifying the trends and keeping narrow prediction intervals. 
Overall, the normalizing flow yields the sharpest prediction intervals, while the Gaussian copula shows low sharpness.

On the first day shown in Figure~\ref{fig:Price_Delta_Example_Prediction_Intervals}, there is a significant peak in the ID$_3$ price realization. 
This type of extreme event may occur due to unforeseen incidences in the markets, such as unplanned changes in the operation of large consumers or suppliers. 
Notably, the normalizing flow is the only approach that captures this peak. 
The supplementary material shows the prediction intervals for other peak days in the test month that confirm the normalizing flows' ability to capture the price peaks.

\begin{table}
\centering
\caption{Percentage of realizations within 50\% and 90\% prediction intervals (PI). }
\label{tab:Price_Delta_Prediction_Interval_Coverage}
\begin{tabular}{@{}lll@{}}
\toprule
                 & PI 50\%  & PI 90\%  \\ \midrule
Normalizing Flow & 59.4\% & 93.5\% \\
Historical       & 62.1\% & 95.5\% \\
Copula           & 61.6\% & 96.0\% \\
Gaussian         & 54.9\% & 90.2\% \\ \bottomrule
\end{tabular}\end{table}
Table~\ref{tab:Price_Delta_Prediction_Interval_Coverage} shows the percentage of time steps in the test set that lie within the 50\% and 90\% prediction interval, respectively. 
All approaches show conservative results for both prediction intervals, i.e., the actual percentage of realizations within the prediction intervals is higher than the selected probability.
Thus, all approaches yield reliable prediction intervals. 
The supplementary material also lists the percentage of realizations within the prediction intervals for each of the four time steps of the price difference vector. The results show that the multivariate approach is able to predict all four dimensions well and that there is no deterioration of the prediction accuracy for the later time steps.

In summary, the normalizing flow shows the best results with sharp prediction interval presented in Figure~\ref{fig:Price_Delta_Example_Prediction_Intervals} and reliable results in Table~\ref{tab:Price_Delta_Prediction_Interval_Coverage}.
The selection of historical data shows surprisingly good results but has poor sharpness in the prediction intervals due to the missing conditional input information. 
The Gaussian copula results in prediction intervals with poor sharpness and extreme peaks that are likely a result of a poor quantile regression of the tails of the price difference distribution. By overestimating the tails in the inverse CDF, the Gaussian copula samples a higher share of outliers than the actual distribution. The high reliability indicated in Table~\ref{tab:Price_Delta_Prediction_Interval_Coverage} comes at the cost of poor sharpness.
Finally, the Gaussian regression predicts 14 values for the mean vector and a lower-triangular covariance matrix \citep{dillon2017tensorflow}. Likely, this number is too high to achieve good fits in the regression model, and instead, the training regresses to constant outputs.

\subsection{Energy and variogram score}\label{sec:Price_Delta_Energy_Variogramm}
\begin{figure*}
    \centering
    \includegraphics[width=\textwidth]{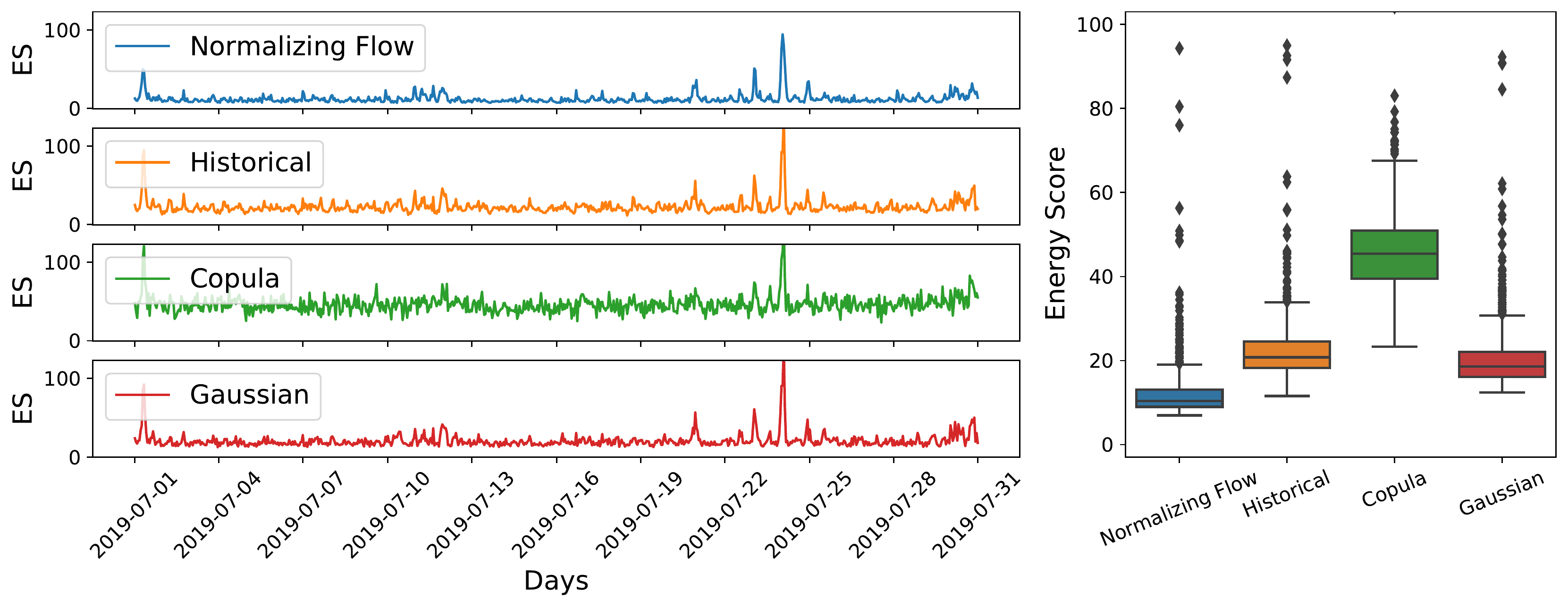}
    \caption{Energy score \citep{gneiting2008assessing,pinson2012evaluating} of the normalizing flow (``Normalizing Flow''), the selection of historical data (``Historical''), the Gaussian copula (``Copula''), and the Gaussian regression (``Gaussian'') for each hour in test month of July 2019 (left) and box plot \citep{waskom2021seaborn} of the overall distribution for each model (right). 
    Training data from January 2018 to June 2019 \citep{EnergyCharts2022}.
    Conditional inputs as derived in Section~\ref{sec:Price_Delta_DerivingFeatureVector}.
    }
    \label{fig:Price_Delta_EnergyScore}
\end{figure*}

Figure~\ref{fig:Price_Delta_EnergyScore} shows the energy score for each model and each hour of the test month (left), as well as box plots of the overall energy score distributions (right). 
The Gaussian regression performs only marginally better than the historical selection. 
As shown in Figure~\ref{fig:Price_Delta_Example_Prediction_Intervals}, the Gaussian regression struggles to identify the correct trends, and the historical selection shows wider prediction intervals, i.e., both approaches show weaknesses that lead to an increase in the energy score in different ways.
The Gaussian copula shows significantly worse results than the other forecasting models. As already observed in Figure~\ref{fig:Price_Delta_Example_Prediction_Intervals}, the Gaussian copula results in extensive prediction intervals and fails to identify the intraday price patterns. Thus, the large energy scores in Figure~\ref{fig:Price_Delta_EnergyScore} confirm the qualitative observations from Figure~\ref{fig:Price_Delta_Example_Prediction_Intervals}.
Finally, the normalizing flow shows the lowest energy score, i.e., it provides the most reliable predictions while maintaining sharp prediction intervals with a diverse set of samples.

In general, the energy scores over time in the case of all methods show rare incidences of extreme price peaks, e.g., on the first day and at the beginning of the last quarter of the month. 
Notably, however, the normalizing flow shows the best approximation of these extreme events, e.g., for the price peak on the first day of the test month shown in Figure~\ref{fig:Price_Delta_Example_Prediction_Intervals} and the other peak days shown in the supplementary material.

\begin{figure*}
    \centering
    \includegraphics[width=\textwidth]{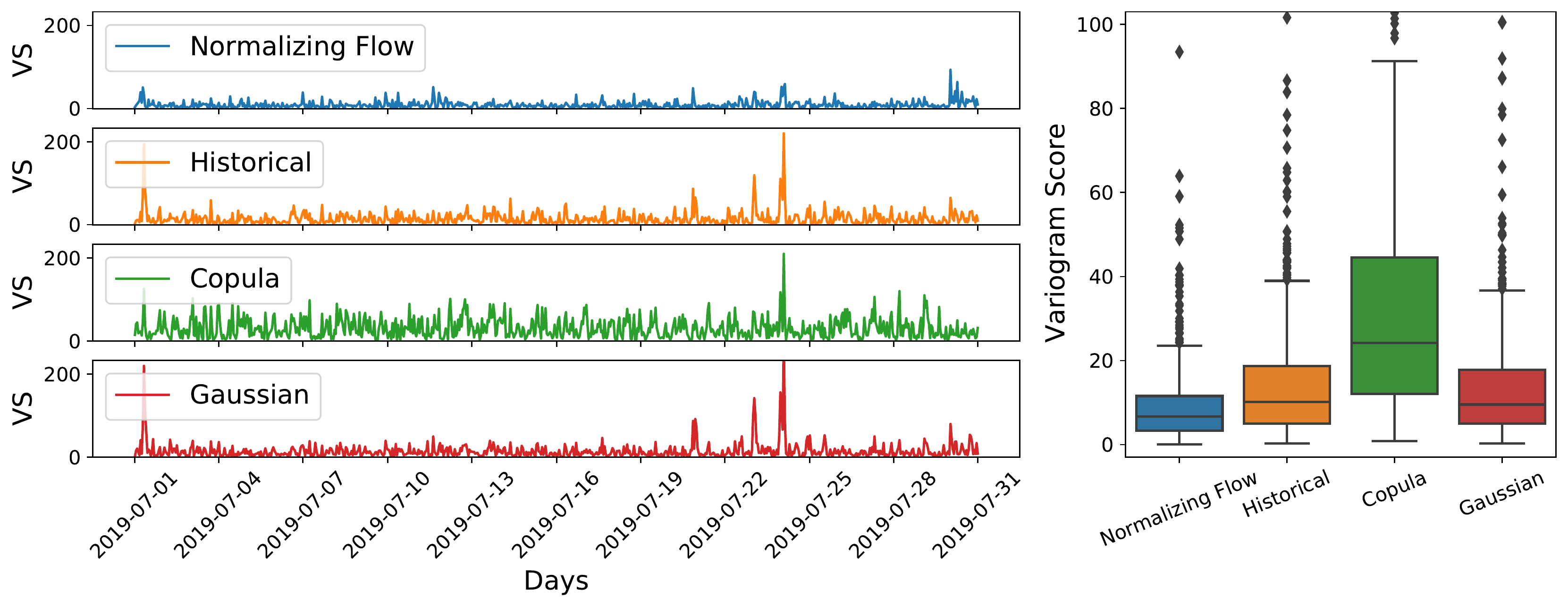}
    \caption{Variogram score \citep{scheuerer2015variogram} of the normalizing flow (``Normalizing Flow''), the selection of historical data (``Historical''), the Gaussian copula (``Copula''), and the Gaussian regression (``Gaussian'') for each hour in test month of July 2019 (left) and box plot \citep{waskom2021seaborn} of the overall distribution for each model (right). 
    Training data from January 2018 to June 2019 \citep{EnergyCharts2022}.
    Conditional inputs as derived in Section~\ref{sec:Price_Delta_DerivingFeatureVector}.
    }
    \label{fig:Price_Delta_VariogramScore}
\end{figure*}
Next, the variogram score is used to compare whether the different models can identify the correlation between the dimensions. Figure~\ref{fig:Price_Delta_VariogramScore} shows the variogram score for each model and each hour of the test month (left), as well as box plots of the overall distributions (right).
Note that the y-axis of the right of Figure~\ref{fig:Price_Delta_VariogramScore} is truncated to allow for better visualization of the distributions.
Similar to the energy score results, the normalizing flow shows the best results, while the Gaussian regression performs approximately equal to the historical selection. The Gaussian copula appears to miss large parts of the correlation and shows the worst variogram scores. 
The flexibility of the normalizing flow allowed the model to learn the correlation between the four time steps in the price difference vector better than any other considered approach. In particular, the days with extreme price peaks result in consistently low variogram scores for the normalizing flow, while all other models result in spiking variogram scores. 
In the supplementary material, we show the evaluations of energy score and variogram score over a longer period from July to December 2019. The results confirm the observations described here.

In conclusion, the flexibility of the normalizing flow can describe the non-trivial distribution of the considered price differences, as highlighted by both the energy scores and the variogram scores for the test month.
Furthermore, the combination with external impact factors leads to a significant improvement over the historical selection. 
Meanwhile, the Gaussian copula results in erroneous trends and inflated prediction intervals, and the Gaussian regression shows no advantage compared to the selection of historical data.

\subsection{Significance of external factors}\label{sec:Delta_Price_Empirical_Significants_Features}
\begin{figure}
    \centering
    \includegraphics[width=\columnwidth]{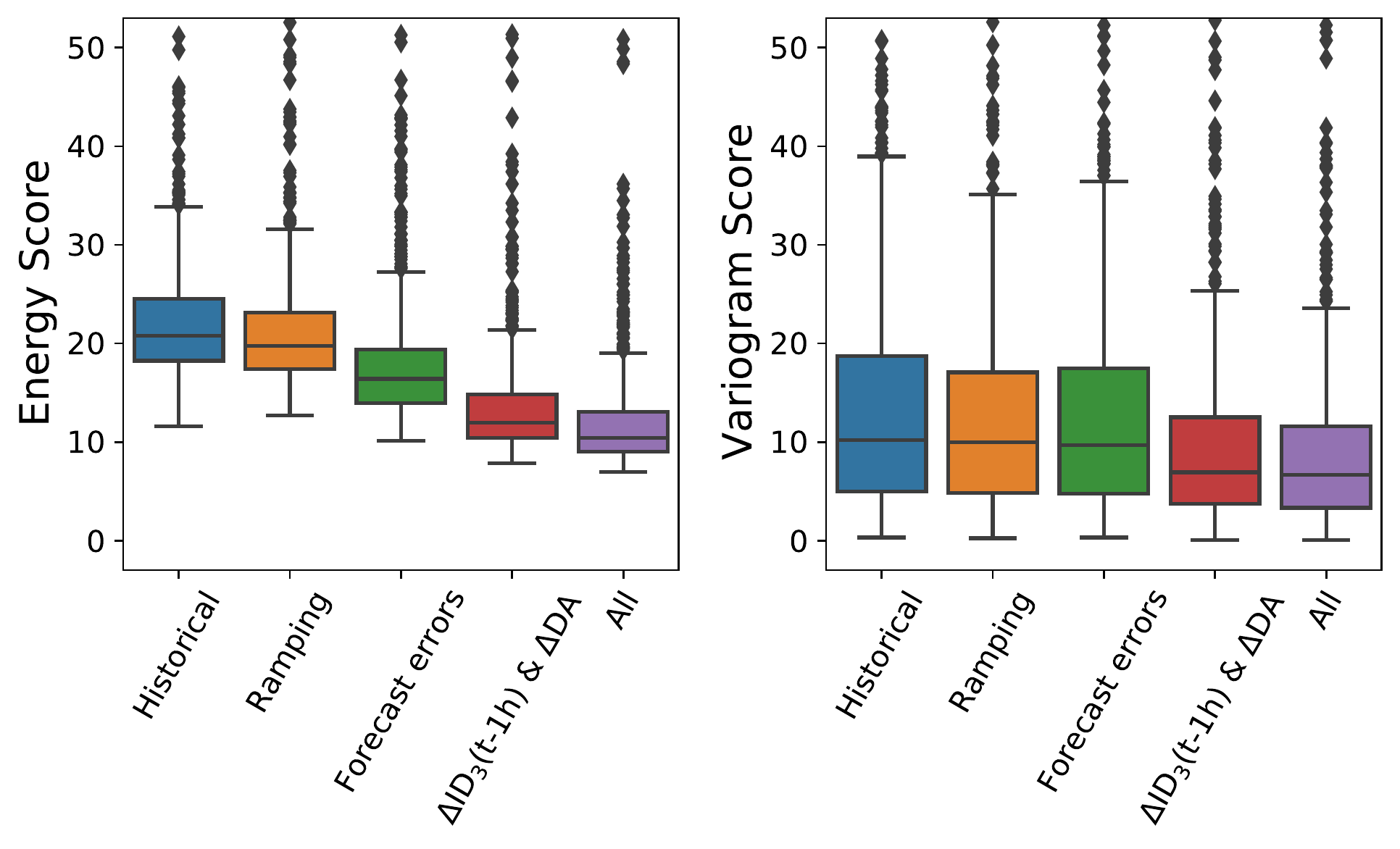}
    \caption{Energy score \citep{gneiting2008assessing,pinson2012evaluating} and variogram score \citep{scheuerer2015variogram} of normalizing flow predictions using different sets of external factors. Note the truncated y-axis. 
    Training data from January 2018 to June 2019 \citep{EnergyCharts2022} and July 2019 as the test set.
    }
    \label{fig:Price_Delta_EnergyScore_InputFactors}
\end{figure}

Up to this point, it remains unclear which external features contribute to the performance of the normalizing flow predictions.
The SHAP-value analysis in Section~\ref{sec:Delta_Price_XAI} has provided first insights, albeit for a different deterministic prediction model. 
This Subsection now compares the performance of the normalizing flow using different sets of input features. 
Figure~\ref{fig:Price_Delta_EnergyScore_InputFactors} shows box plots of the energy score and variogram score distributions over the test month for samples generated using the normalizing flow with the different groups of input features in comparison to the selection of historical data. Note the truncated y-axis to highlight the majority parts of the distributions. 

The energy score results confirm the observations from the SHAP-value analysis in Section~\ref{sec:Delta_Price_XAI}. 
The combination of the previous realization of the price difference vector and the day-ahead increments (``$\bm{\Delta} \textbf{ID}_3$(t-1h) \& $\Delta\text{DA}$'') results in the largest improvement. 
The forecast errors (``Forecast errors'') lead to a moderate improvement, while the ramping features (``Ramping'') only result in a minor improvement of the energy score compared to the historical selection (``Historical'').
Notably, the combination of all external factors (``All'') achieves an energy score marginally below the combination of the previous realization of the price difference vector and the day-ahead increments.

The variogram score confirms the observations from the energy score. 
Except for the combination of all external factors (``All'') and the combination of the previous realization of the price difference vector and the day-ahead increments (``$\bm{\Delta} \textbf{ID}_3$(t-1h) \& $\Delta\text{DA}$''), all inputs lead to approximately equal results compared to the informed historical selection. 

In conclusion, the combination of the previous realization of the price difference vector and the day-ahead increments has by far the strongest impact on the performance of the probabilistic forecasts and appears to be sufficient to achieve high-quality prediction results. 
However, the normalizing flow can still extract additional information from the remaining external factors.

\cite{narajewski2020econometric} and \cite{han2022complexity} argue that the intraday markets are not efficient in the sense of the efficient-market hypothesis as there are strong relations between the current and previous realizations.  
The results of this work confirm this observation.

\subsection{Discussion}
The analysis in this Section shows promising results that indicate valuable density forecasts of the price difference vector by the normalizing flow using appropriate external input features. Both the Gaussian copula and the Gaussian regression show weaknesses and fail to improve on the results of the informed historical selection suggesting they are not suitable prediction methods.
Figure~\ref{fig:Price_Delta_EnergyScore_InputFactors} shows that only the knowledge of the realizations in the previous hour and the day-ahead increments offers a significant advantage over using the informed selection of historical data. Hence, lead times of multiple hours rescind the advantage of the normalizing flow over the informed historical selection. 
In conclusion, the normalizing flow yields high-quality density forecasts for short lead times. If the available information is limited, the selection of historical price differences based on the hour of the day appears to be sufficient.    \section{Conclusion}\label{sec:Price_Delta_Conclusion}
This work proposes a modeling approach for probabilistic forecasting of intraday electricity prices in the German EPEX spot market. 
The novel approach captures the strong relation between day-ahead and intraday prices by assuming fixed day-ahead prices and predicting the difference between the two price time series. 
Furthermore, the model captures the hourly patterns of fluctuation of the intraday prices around the day-ahead prices by considering each day-ahead trading interval as a four-dimensional joint distribution.

Normalizing flow, a non-parametric distribution model, is used to learn the proposed four-dimensional distribution of price differences and includes external input factors to form a conditional density forecast. 
The proposed model yields a complete end-to-end approach for multivariate density estimation and probabilistic regression that does not require any a priori assumptions about the data.

Compared to a selection of historical data and other multivariate probabilistic models, namely the Gaussian copula and multivariate Gaussian regression, the normalizing flow returns the most reliable and sharpest predictions. 
Using methods from XAI and careful performance analysis, the role of different input features has been analyzed in detail. 
In summary, the normalizing flow best identifies the trends of the intraday prices and shows the narrowest prediction intervals.

\section*{Declaration of Competing Interest}
We have no conflict of interest.

\section*{Acknowledgements}
We gratefully acknowledge Johannes Kruse (FZ-J\"ulich IEK-STE), Julius Trebbien (FZ-J\"ulich IEK-STE), and Leonard Rydin Gorj\~ao (Department of Computer Science, OsloMet) for the discussion about the concepts of explainable AI and the importance of different impact factors. 
This work was performed as part of the Helmholtz School for Data Science in Life, Earth and Energy (HDS-LEE) and received funding from the Helmholtz Association of German Research Centres.

 \section*{Nomenclature}\label{sec:Price_Delta_Nomenclature}

\noindent\begin{tabularx}{\columnwidth}{XX}
    Symbol                      & Description \\
    \hline
    $f(\mathbf{y})$                      & Deterministic forecasting model \\ 
    FI                          & Feature importance \\
    $N$                         & Number of samples \\
$P^{RE}_t$                  & Power values $RE \in \{\text{Solar, Wind, Load} \}$\\
    $\hat{P}^{RE}_t$            & Forecast of renewable electricity \\
    $\Delta P^{RE}_t$           & Forecast error of renewable electricity\\
    $T$                         & Invertible neural network \\ 
    $\mathbf{J}_{T^{-1}}$       & Jacobian of $T^{-1}$ \\
    $\hat{T}$                   & Day-ahead trading interval $\hat{T}=4$ \\
    $x_t$                       & Value at time step $t$ \\
    $\hat{x}_t$                 & Sample at time step $t$ \\
    $\mathbf{x}, \mathbf{y}$    & Data vector \\
    $\hat{\mathbf{z}}$          & Gaussian sample \\ 
    $\hat{\mathbf{x}}, \hat{\mathbf{y}}$ & Sample of data vector \\
    $\mathbf{y}$                & External input factors\\
    $\text{ID}_3$               & Intraday price \\
    $\text{ID}^{00}_3$, $\text{ID}^{15}_3$, $\text{ID}^{30}_3$, $\text{ID}^{45}_3$  & Quater hour steps of intraday prices\\
    $\text{DA}$                 & Day-ahead price\\
    $\bm{\Delta}\textbf{ID}_3 $ & 4D vector of price difference\\
    $\phi$                      & Multivariate standard Gaussian \\ 
    $\varphi$                   & SHAP value \\ 
    $\mathcal{N}_{X\vert Y}\left(\mathbf{x}; \bm{\mu}_X(\mathbf{y}), \bm{\Sigma}_X(\mathbf{y}) \right)$ & Conditional Gaussian distribution \\
    $p_{X\vert Y}(\mathbf{x}\vert \mathbf{y})$        & Conditional PDF of $X$ given $Y$ \\
    $p_{\bm{\Delta}\textbf{ID}_3\vert Y}(\bm{\Delta}\textbf{ID}_3 \vert \mathbf{y})$        & Conditional PDF of $\bm{\Delta}\textbf{ID}_3$ given $Y$ \\

$\mathcal{N}_{X\vert Y}$    & Gaussian distribution with moments predicted via inputs Y \\
    $\bm{\mu}_X(\mathbf{y})$    & Neural network mean value predictor \\
    $\bm{\Sigma}_X(\mathbf{y})$ & Neural network covariance matrix value predictor \\
    $\gamma$                    & Variogram order (typically $\gamma = 0.5$) \\
\end{tabularx}

  \appendix

\bibliographystyle{apalike}
  \renewcommand{\refname}{Bibliography}

  \bibliography{extracted.bib}

\end{document}


\ifx\REVIEW\undefined
\twocolumn[
\begin{@twocolumnfalse}
\fi
  \thispagestyle{firststyle}

  \begin{center}
    \begin{large}
      \textbf{\mytitle}
    \end{large} \\
    \myauthor
  \end{center}

  \vspace{0.5cm}

  \begin{footnotesize}
    \affil
  \end{footnotesize}

  \vspace{0.5cm}
\ifx\REVIEW\undefined
\end{@twocolumnfalse}
]
\fi

\section{Input feature selection}
Table~\ref{tab:Price_PearsonInputFactors_Price}, Table~\ref{tab:Price_PearsonInputFactors_Error}, and Table~\ref{tab:Price_PearsonInputFactors_Ramps} list the Pearson correlation of all considered external factors with the 4 dimensions of the price difference vector $\bm{\Delta}\textbf{ID}_3$. 
The Pearson correlation indicates the highest dependency on the immediate history of the price difference vector.

\begin{table*}
    \centering
    \caption{Pearson correlation between the price difference dimensions $\bm{\Delta}\textbf{ID}_3$ and the price and time features. 
    00, \dots, 45 indicate the for 15\,min intervals in each day-ahead trading interval.
    Price data from \cite{EnergyCharts2022} and renewables data from ENTSO-E \citep{entsoe2022}. }
    \label{tab:Price_PearsonInputFactors_Price}
\begin{tabular}{@{}lllll@{}}
        \toprule
        External factors          & $\Delta\text{ID}_3^{00}$ & $\Delta\text{ID}_3^{15}$ & $\Delta\text{ID}_3^{30}$ & $\Delta\text{ID}_3^{45}$ \\ \midrule
        DA                       & -0.02                  & -0.04                  & -0.05                  & -0.04                  \\
        $\Delta\text{DA}^{-}$     & -0.44                  & -0.16                  & 0.14                   & 0.31                   \\
        $\Delta\text{DA}^{+}$     & 0.38                   & 0.10                   & -0.22                  & -0.43                  \\
        cos(t)                    & 0.11                   & 0.03                   & -0.04                  & -0.10                  \\
        sin(t)                    & 0.01                   & 0.00                   & 0.02                   & -0.01                  \\
        $\Delta\text{ID}_3^{00} (t-1h)$ & 0.72                   & 0.55                   & 0.28                   & 0.05                   \\
        $\Delta\text{ID}_3^{15} (t-1h)$ & 0.70                   & 0.69                   & 0.51                   & 0.31                   \\
        $\Delta\text{ID}_3^{30} (t-1h)$ & 0.49                   & 0.67                   & 0.69                   & 0.56                   \\
        $\Delta\text{ID}_3^{45} (t-1h)$ & 0.24                   & 0.53                   & 0.70                   & 0.67                   \\
        $\Delta\text{ID}_3^{00} (t-2h)$ & 0.46                   & 0.37                   & 0.22                   & 0.09                   \\
        $\Delta\text{ID}_3^{15} (t-2h)$ & 0.51                   & 0.50                   & 0.38                   & 0.24                   \\
        $\Delta\text{ID}_3^{30} (t-2h)$ & 0.44                   & 0.52                   & 0.48                   & 0.36                   \\
        $\Delta\text{ID}_3^{45} (t-2h)$ & 0.31                   & 0.45                   & 0.48                   & 0.42                   \\
        \bottomrule
    \end{tabular}\end{table*}

\begin{table*}
    \centering
    \caption{Pearson correlation between the price difference dimensions $\bm{\Delta}\textbf{ID}_3$ and the forecast errors. 
    00, \dots, 45 indicate the for 15\,min intervals in each day-ahead trading interval.
    Price data from \cite{EnergyCharts2022} and renewables data from ENTSO-E \citep{entsoe2022}. }
    \label{tab:Price_PearsonInputFactors_Error}
\begin{tabular}{@{}lllll@{}}
        \toprule
        External factors          & $\Delta\text{ID}_3^{00}$ & $\Delta\text{ID}_3^{15}$ & $\Delta\text{ID}_3^{30}$ & $\Delta\text{ID}_3^{45}$ \\ \midrule
        Solar Error 00            & -0.20                  & -0.23                  & -0.21                  & -0.15                  \\
        Solar Error 15            & -0.20                  & -0.24                  & -0.23                  & -0.17                  \\
        Solar Error 30            & -0.20                  & -0.24                  & -0.24                  & -0.18                  \\
        Solar Error 45            & -0.20                  & -0.24                  & -0.24                  & -0.19                  \\
        Wind Error 00             & -0.28                  & -0.36                  & -0.38                  & -0.32                  \\
        Wind Error 15             & -0.28                  & -0.36                  & -0.39                  & -0.33                  \\
        Wind Error 30             & -0.28                  & -0.37                  & -0.39                  & -0.33                  \\
        Wind Error 45             & -0.28                  & -0.36                  & -0.39                  & -0.34                  \\
        Load Error 00             & 0.12                   & 0.11                   & 0.09                   & 0.04                   \\
        Load Error 15             & 0.12                   & 0.11                   & 0.09                   & 0.04                   \\
        Load Error 30             & 0.12                   & 0.11                   & 0.09                   & 0.04                   \\
        Load Error 45             & 0.12                   & 0.11                   & 0.08                   & 0.03                   \\
        \bottomrule
    \end{tabular}\end{table*}  
\begin{table*}
    \centering
    \caption{Pearson correlation between the price difference dimensions $\bm{\Delta}\textbf{ID}_3$ and the ramping features. 
    00, \dots, 45 indicate the for 15\,min intervals in each day-ahead trading interval.
    Price data from \cite{EnergyCharts2022} and renewables data from ENTSO-E \citep{entsoe2022}. }
    \label{tab:Price_PearsonInputFactors_Ramps}
\begin{tabular}{@{}lllll@{}}
        \toprule
        External factors          & $\Delta\text{ID}_3^{00}$ & $\Delta\text{ID}_3^{15}$ & $\Delta\text{ID}_3^{30}$ & $\Delta\text{ID}_3^{45}$ \\ \midrule
        DA solar ramp           & 0.23                   & 0.08                   & -0.07                  & -0.21                  \\
        DA wind ramp            & 0.05                   & -0.03                  & -0.10                  & -0.14                  \\
        DA load ramp            & -0.27                  & -0.09                  & 0.12                   & 0.25                   \\
        DA total gen ramp       & -0.02                  & -0.01                  & 0.03                   & -0.01                  \\
        DA import/export ramp   & -0.27                  & -0.09                  & 0.09                   & 0.26                   \\
        Oil price                 & -0.01                  & 0.01                   & 0.01                   & 0.00                   \\
        Gas price                 & -0.02                  & -0.01                  & -0.02                  & -0.03                  \\
        \bottomrule
    \end{tabular}\end{table*}

 \section{Time dependency}
\begin{figure*}
    \centering
    \includegraphics[width=\textwidth]{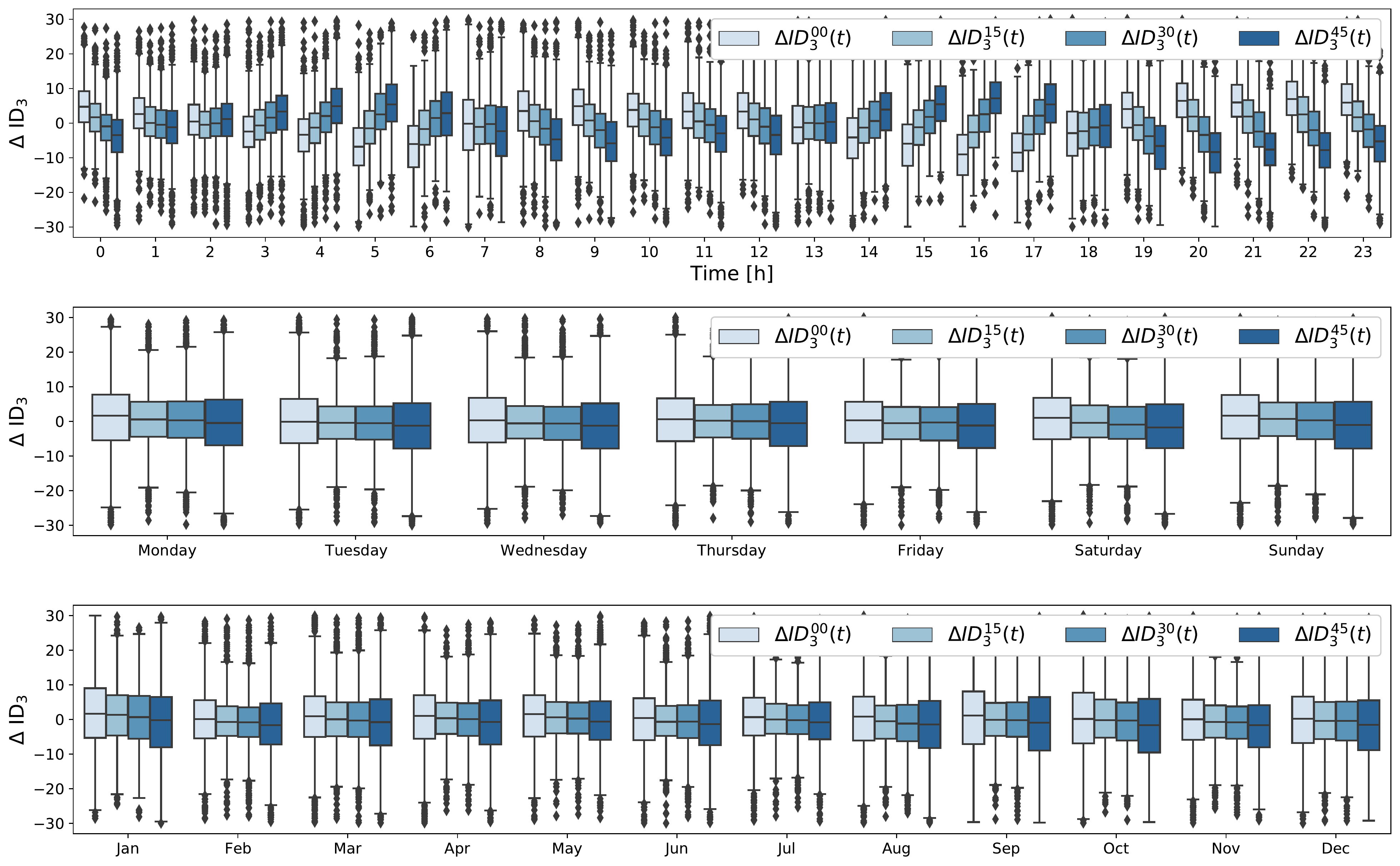}
    \caption{Box plots of distribution of price difference as function of hour of the day, day of the week, and month of the year.}
    \label{fig:Price_Delta_SI_TimeDependency}
\end{figure*}

Figures~\ref{fig:Price_Delta_SI_TimeDependency} shows the dependency of the price difference vector on the hour of the day, the day of the week, and the month of the year, respectively. 
Notably, only the hour of the day appears to have a significant impact on the realizations. This also indicates a similar behavior of the price differences for weekdays and weekends. 

 \section{Shapley Additive exPlanations}

For black-box machine learning models, the quantitative relationship between input features and model outputs is often unknown. Here, explainable artificial intelligence (XAI) offers methods to estimate the quantitative relationship between features and outputs. 
A simple approach estimates the impact of different features using variations under the ceteris paribus assumption, i.e., by varying a single feature while keeping the remaining features constant. 
However, such one-at-a-time variations ignore possible combinatorial effects, which often lead to nonlinear behavior in the outputs. 
Shapley Additive exPlanations (SHAP) values are designed to describe the impacts of the input features on the model output including nonlinear combinatorial effects \citep{lundberg2017unified}. 
Thus, SHAP values give quantitative insight into the underlying systematic behavior of the system and can, e.g., be used to reveal the driving factors of a stochastic process like the power grid frequency \citep{kruse2021revealing} or electricity prices \citep{trebbien2022understanding}.

SHAP values are computed by combining the marginal feature perturbations, i.e., the impact of including or excluding the feature from the feature set, for \textit{all} possible coalitions and combinations of features. 
Notably, this evaluation of all possible combinations is computationally prohibitively expensive for large feature sets and arbitrary models. However, there exist efficient algorithms that compute SHAP values for tree-based models in polynomial time \citep{lundberg2017unified,lundberg2020local}. For more information on the computation of SHAP values, the reader is referred to the original publications \citep{lundberg2017unified,lundberg2020local} and the documentation of the SHAP python package \citep{lundberg2017unified}.

SHAP values are based on the idea to express a model output $f(\mathbf{y})$ using an additive feature attribution \citep{lundberg2020local}:
\begin{equation}\label{Eq:SI_Additive_feature_attribution}
    f(\mathbf{y}) \approx \varphi_0 + \sum_i  \varphi_i(y_i;f)
\end{equation}
Here, $\varphi_0$ is the null output, and $\varphi_i(y_i;f)$ are the SHAP values for each input feature $i$. The null output is computed as the expected model output over the full dataset: 
\begin{equation*}
    \varphi_0 = \frac{1}{N} \sum_t f(\mathbf{y}_t)
\end{equation*}
Here, $N$ is the number of points in the dataset. 
For tree-based models, like the ones used for the SHAP analysis in the main paper, the SHAP algorithm provides exact local accuracy, i.e., the approximate additive feature attribution in Equation~\eqref{Eq:SI_Additive_feature_attribution} is met exactly. 
The formulation via the additive feature attribution also ensures the \textit{missingness} in the model such that, if a SHAP value is zero, the model output does not change. 
Furthermore, the SHAP value decomposition is consistent, i.e., a change to an input feature leads to a change in the corresponding SHAP value \citep{lundberg2017unified}.

 \section{Additional analysis of Winkler score}
The Winkler score \citep{winkler1972decision, gneiting2007strictly} is a well-established test to investigate reliability and sharpness:
\begin{equation*}
    \text{WS} = 
        \begin{cases}
            \delta_t  & x_t \in \left[ \hat{L}^\alpha_t, \hat{U}^\alpha_t \right]  \\
            \delta_t + \frac{2}{1-\alpha} \left(\hat{U}^\alpha_t - x_t \right) & x_t \geq  \hat{U}^\alpha_t \\
            \delta_t + \frac{2}{1-\alpha} \left(x_t - \hat{L}^\alpha_t \right) & x_t \leq  \hat{L}^\alpha_t 
        \end{cases}
\end{equation*}
Here, $x_t$ is the time step, $\hat{L}^\alpha_t$ and $\hat{U}^\alpha_t$ are the lower and upper bound of the confidence interval with probability $\alpha$, respectively. The term $\delta_t$ indicates the width of the predicted confidence interval:
\begin{equation*}
    \delta_t = \hat{U}^\alpha_t - \hat{L}^\alpha_t
\end{equation*}
The Winkler score is a negatively oriented score, i.e., lower values indicate better performance.

\begin{figure*}
    \centering
    \includegraphics[width=0.7\textwidth]{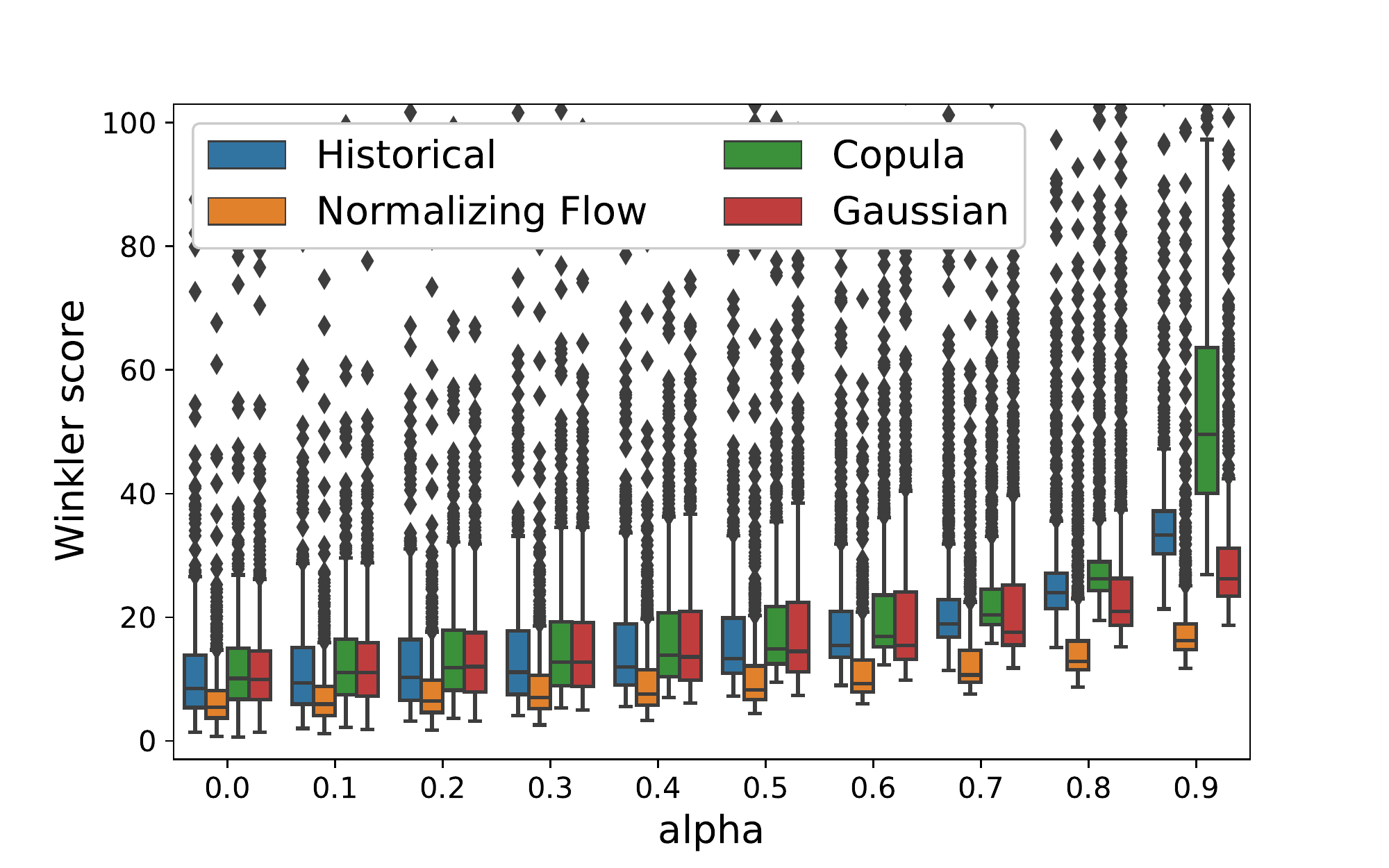}
    \caption{Winkler score \citep{winkler1972decision, gneiting2007strictly} over confidenc interval width \textit{alpha} for the different models and the historical data. 
    Training data from January 2018 to June 2019.
    Full distribution of scores over the test month of July 2019. }
    \label{fig:Price_Delta_WinklerScore}
\end{figure*}

Figure~\ref{fig:Price_Delta_WinklerScore} shows the distributions of Winkler scores over the full month of July 2019 for each of the probabilistic models and over $\alpha$ values between 0.0 and 0.9. Overall, the normalizing flow shows the best results. 

 \section{ANN structures}
The Gaussian regression uses a fully-connected neural network with one hidden layer, 32 neurons, and `tanh' activation. The output layer has 14 neurons to describe the four-dimensional mean and the lower triangular covariance matrix implemented in TensorFlow-probability \citep{dillon2017tensorflow}. The neuron that output the mean have no activation and the neurons that output the covariance matrix use `softmax' activation to ensure strict positivity. 

The normalizing flow uses the affine coupling layer implementation in \citep{dinh2016realNVP}. 
The full model uses four affine coupling layers with fully connected conditioner models. The conditioner models have two hidden layers with `ReLU' activation and four neurons per layer. 
 \section{Prediction intervals on extreme peaks}
\begin{figure*}
    \centering
    \includegraphics[width=0.93\textwidth]{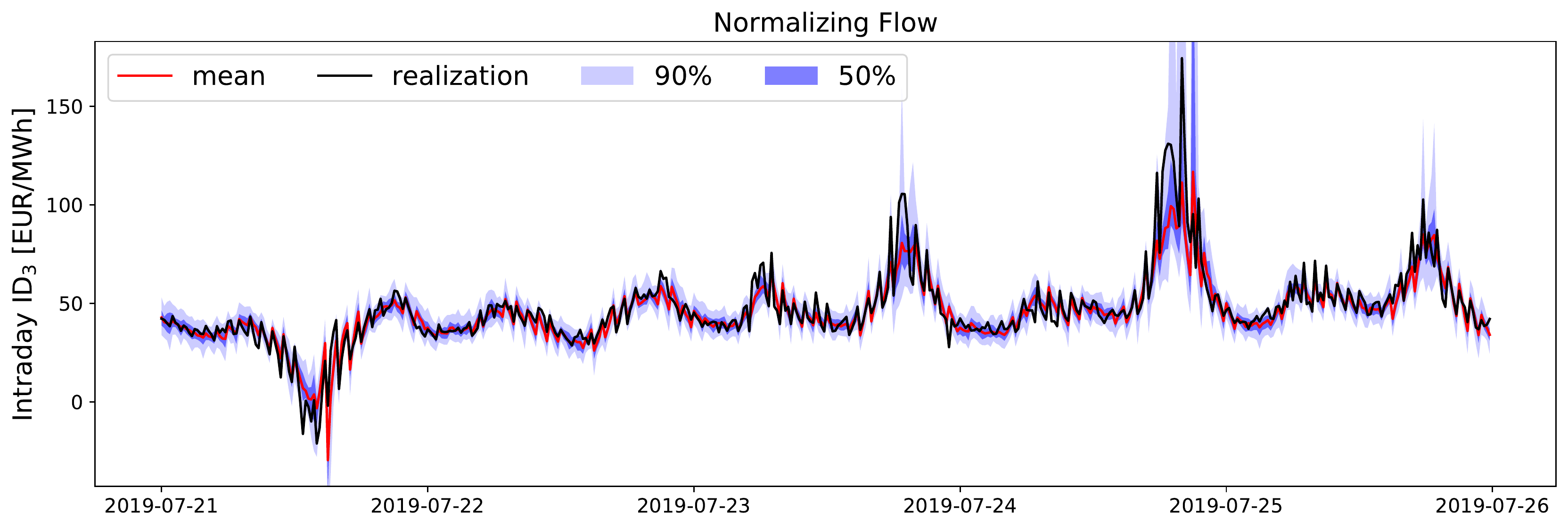}
    \includegraphics[width=0.93\textwidth]{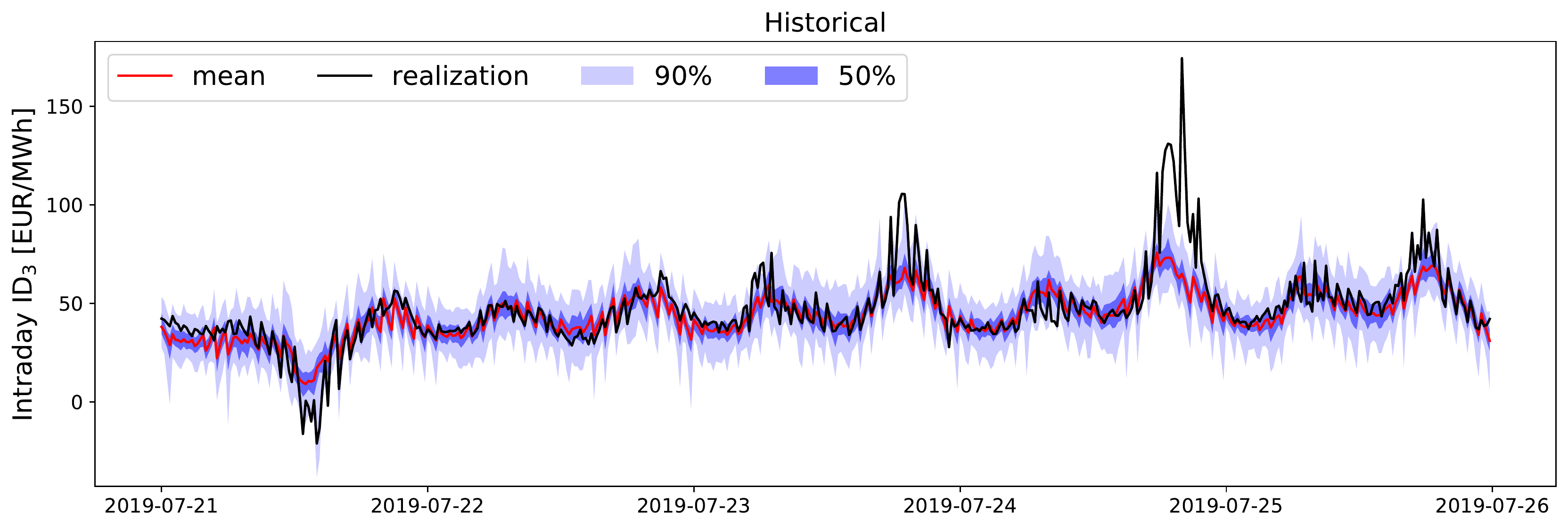}
    \includegraphics[width=0.93\textwidth]{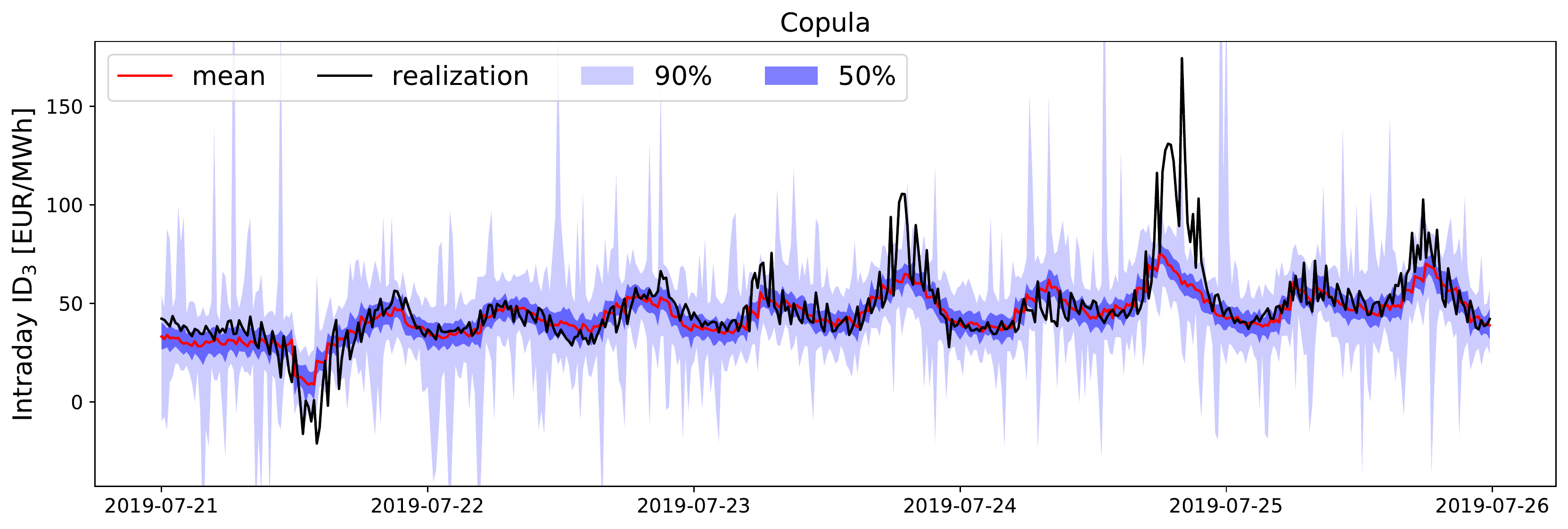}
    \includegraphics[width=0.93\textwidth]{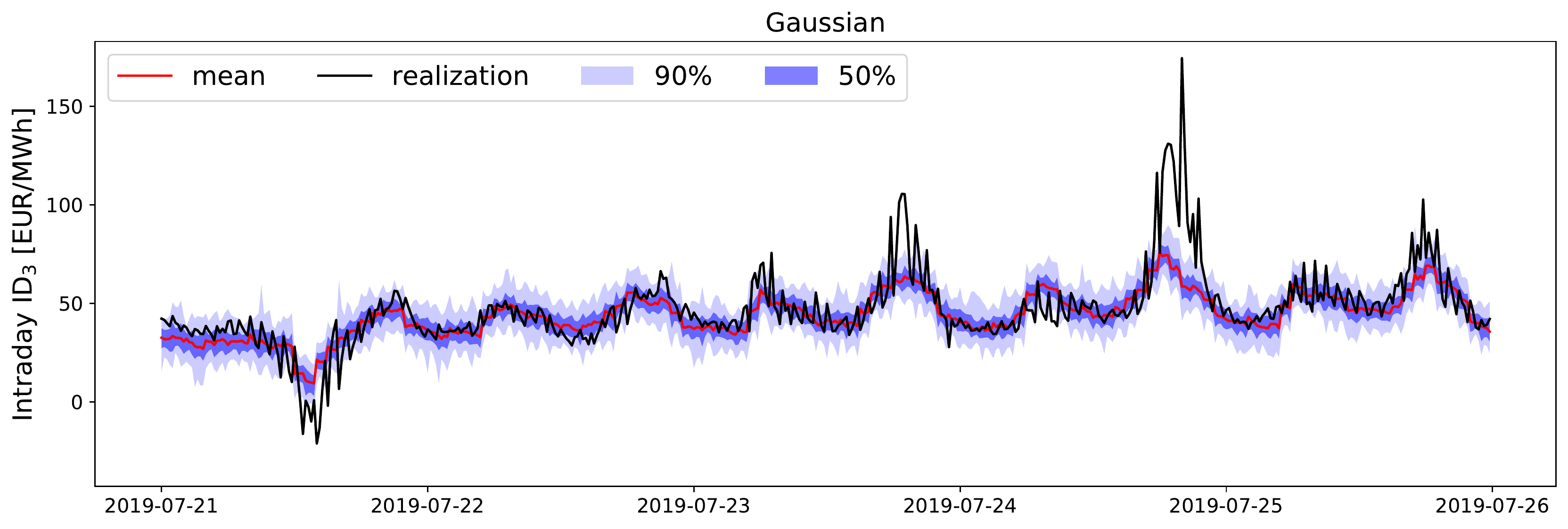}
    \caption{Predicted mean, prediction intervals of 50\% and 90\%, and realization between July 21$^{\text{st}}$ and July 26$^{\text{th}}$ of 2019 estimated from the probabilistic forecasts form the normalizing flow (top), the selection of historical data (second from top), the Gaussian copula (third from top), and the Gaussian regression (bottom). 
    Note the truncated y-axis for the Gaussian copula.
    Training data from January 2018 to June 2019 \citep{EnergyCharts2022}.
    Results generated with all conditional inputs.}
    \label{fig:Price_Delta_SI_Example_Confidence_Intervals_Input_Factors}
\end{figure*}

Figure~\ref{fig:Price_Delta_SI_Example_Confidence_Intervals_Input_Factors} shows the predicted mean, the prediction intervals of 50\% and 90\%, and the realization between July 21$^{\text{st}}$ and July 26$^{\text{th}}$ of 2019. Similar to the prediction intervals shown in the main part of the paper, the results confirm that the normalizing flow results in the narrowest prediction intervals and is the only approach that captures the extreme price peaks. 

 \section{Quarter-hourly resolution of prediction interval accuracy}
Table~\ref{tab:Price_Delta_Prediction_Interval_Coverage_quater} shows the share of realizations within the prediction intervals for each of the four dimensions of the distribution predicted by the normalizing flow. The results show no deterioration of performance for the later time steps. 
Thus, any concern about limitations of the multivariate model being applied to the step-by-step realizing intraday price are theoretical rather than practical. 

\begin{table}
\centering
\caption{Percentage of realizations within 50\% and 90\% prediction intervals (PI) for each quarter hour. Results for normalizing flow over test interval from July 1$^{\text{st}}$, 2019 to December 31$^{\text{st}}$, 2019. }
\label{tab:Price_Delta_Prediction_Interval_Coverage_quater}
\begin{tabular}{@{}lllll@{}}
\toprule
                & $\Delta\text{ID}_3^{00}$ & $\Delta\text{ID}_3^{15}$ & $\Delta\text{ID}_3^{30}$ & $\Delta\text{ID}_3^{45}$  \\ \midrule
PI 50\%         & 0.54      & 0.53      & 0.54      & 0.62 \\ 
PI 90\%         & 0.91      & 0.90      & 0.91      & 0.93 \\ \bottomrule
\end{tabular}\end{table}

 \section{Energy and variagram score}

\begin{figure*}
    \centering
    \includegraphics[width=\textwidth]{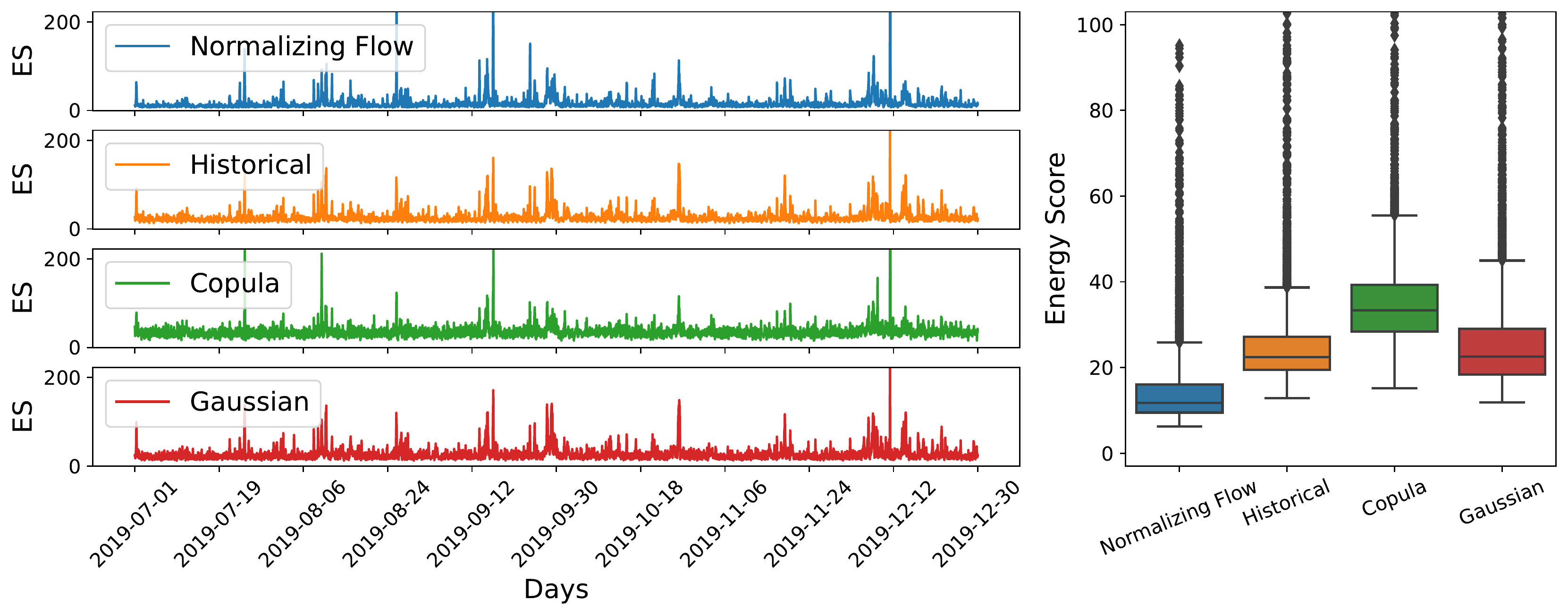}
    \caption{Energy score \citep{gneiting2008assessing,pinson2012evaluating} of the normalizing flow (``Normalizing Flow''), the selection of historical data (``Historical''), the Gaussian copula (``Copula''), and the Gaussian regression (``Gaussian'') for each hour from July to December 2019 (left) and box plot \citep{waskom2021seaborn} of the overall distribution for each model (right). 
    Training data from January 2018 to June 2019 \citep{EnergyCharts2022}.
    }
    \label{fig:Price_Delta_EnergyScore}
\end{figure*}

\begin{figure*}
    \centering
    \includegraphics[width=\textwidth]{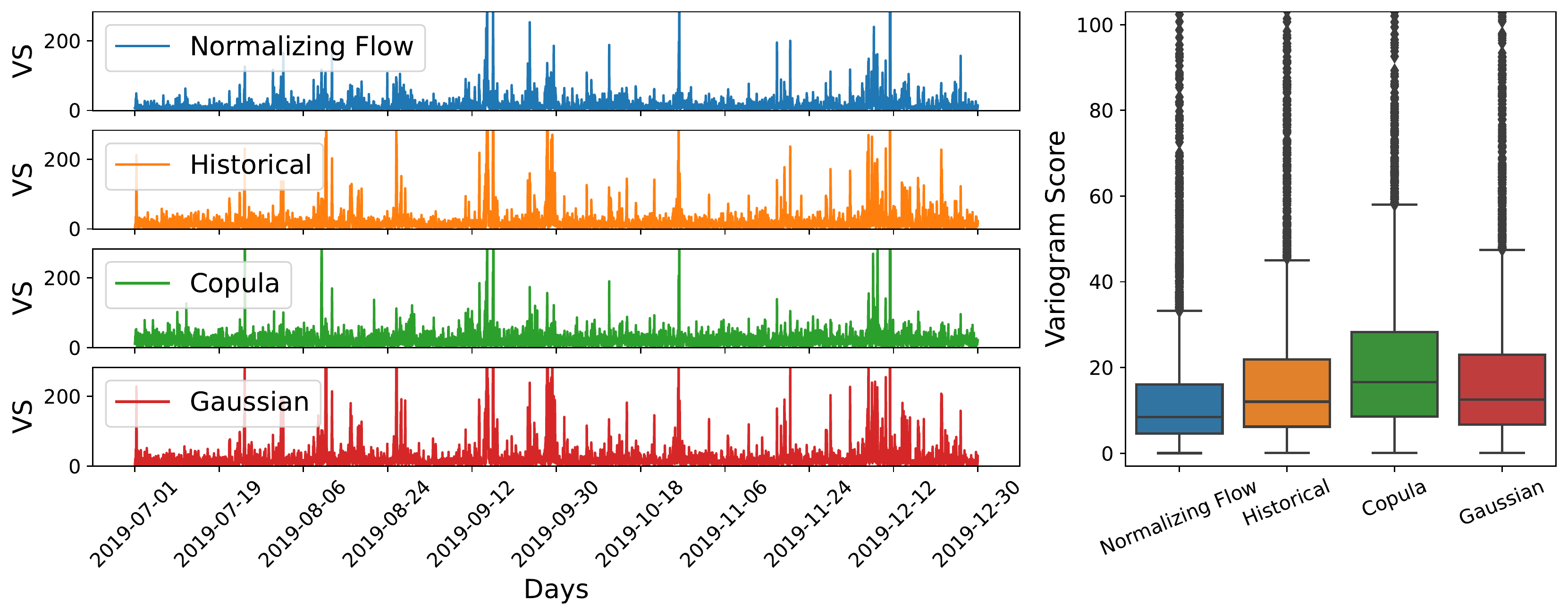}
    \caption{Variogram score \citep{scheuerer2015variogram} of the normalizing flow (``Normalizing Flow''), the selection of historical data (``Historical''), the Gaussian copula (``Copula''), and the Gaussian regression (``Gaussian'') for each hour from July to December 2019 (left) and box plot \citep{waskom2021seaborn} of the overall distribution for each model (right). 
    Training data from January 2018 to June 2019 \citep{EnergyCharts2022}.
    }
    \label{fig:Price_Delta_VariogramScore}
\end{figure*}

Figure~\ref{fig:Price_Delta_EnergyScore} shows the energy score and Figure~\ref{fig:Price_Delta_VariogramScore} shows the variogram score for each model and each hour from July to December 2019 (left), as well as box plots of the overall distributions (right). 
Note that there are some peaks in the results of all methods in both the energy score and the variogramm score, e.g., in September or December. Still, the distribution of the values for energy score and variogram score for the extended test periods confirm the observations from the main paper. 

\newpage
  \appendix

\bibliographystyle{apalike}
  \renewcommand{\refname}{Bibliography}

  \bibliography{extracted.bib}